\documentclass[11pt]{article}
\usepackage{amssymb}
\usepackage{amsthm}
\usepackage{amsmath,amsfonts,amssymb}

\usepackage[mathscr]{eucal}
\usepackage{exscale}
\usepackage{natbib}
\usepackage{bm}
\usepackage{eqlist}
\usepackage[dvipsnames]{color}
\usepackage[Lenny]{fncychap}
\usepackage{multirow}
\usepackage{graphicx}
\usepackage{algpseudocode}
\usepackage{algorithm}
\usepackage{caption}
\usepackage{hyperref}
\usepackage{framed}
\usepackage{color}
\usepackage{booktabs}
\usepackage{makecell}
\usepackage{array}
\usepackage{float}
\usepackage{placeins}
\usepackage{tabularx}
\usepackage{longtable}
\usepackage[table]{xcolor} 

\usepackage{framed}
\usepackage{algpseudocode}

\usepackage{tabu}
\usepackage{tikz}

\usepackage[utf8]{inputenc} 
\usepackage[english]{babel} 
\usepackage[T1]{fontenc}    
\usepackage{amssymb,amsmath,amsfonts} 
\usepackage{braket} 
\usepackage{caption}
\usepackage{algorithm}

\hypersetup{
    colorlinks=true,
    linkcolor=blue,
    citecolor=blue,
    citebordercolor={1 1 0},
}

\renewcommand{\theequation}{\thesection.\arabic{equation}}

\usepackage{authblk}

\title{CMAB: A First National-Scale Multi-Attribute Building Dataset in China Derived from Open Source Data and GeoAI}

\author{
Yecheng Zhang\textsuperscript{a,1},
Huimin Zhao\textsuperscript{a,1},
Ying Long\textsuperscript{b} \thanks{Corresponding author, e-mail: ylong@tsinghua.edu.cn.} \\
\small \textsuperscript{a} \textit{School of Architecture, Tsinghua University, Beijing, 100084, China} \\
\small \textsuperscript{b} \textit{Hang Lung Center for Real Estate, Key Laboratory of Ecological Planning $\&$ Green Building, Ministry of Education, Tsinghua University, Beijing, 100084, China}
}
\date{\today} 

\begin{document}

\maketitle

\footnotetext[1]{Equal Contribution.}

\begin{abstract}
Rapidly acquiring three-dimensional (3D) building data, including geometric attributes like rooftop, height, and orientations, as well as indicative attributes like function, quality, and age, is essential for accurate urban analysis, simulations, and policy updates. Current building datasets suffer from incomplete coverage of building multi-attributes. This paper introduces a geospatial artificial intelligence (GeoAI) framework for large-scale building modeling, presenting the first national-scale Multi-Attribute Building dataset (CMAB), covering 3,667 spatial cities, 29 million buildings, and 21.3 billion m² of rooftops with an F1-Score of 89.93\% in OCRNet-based extraction, totaling 337.7 billion m³ of building stock. We trained bootstrap aggregated XGBoost models with city administrative classifications, incorporating features such as morphology, location, and function. Using multi-source data, including billions of high-resolution Google Earth images and 60 million street view images (SVIs), we generated rooftop, height, function, age, and quality attributes for each building. Accuracy was validated through model benchmarks, existing similar products, and manual SVI validation, mostly above 80\%. Our dataset and results are crucial for global SDGs and urban planning.

\end{abstract}


\renewcommand{\theequation}{\arabic{equation}} 
\setcounter{equation}{0} 
\section{Background \& Summary}\label{secIntro}
Over the past two decades of urbanization, cities worldwide have experienced rapid expansion, with buildings serving as spatial cellular units and cities exhibiting increasingly complex changes in 3D morphology and social functions \citep{BFCY2022, LK2024, LW2024, LQ2024, LX2024, ZH2022}. A thorough understanding of the fine-grained 3D physical and social structures of cities from building datasets has become crucial for urbanization processes, urban energy, urban ecosystems, and government decisions related to carbon emissions and reduction \citep{CY2024a, CZ2024b, WH2024, WX2024, LY2023}. Specifically, building attributes can be categorized into two primary types: geometric attributes, such as building footprints, heights, and orientations; and indicative attributes, which encompass building functions, ages, quality, and vacancy \citep{DN2010, RM2018, CP2013}. Geometric attributes are essential for analyzing urban physical structures and planning city spaces, while indicative attributes are key to understanding the performance and longevity of structures \citep{DK2014, FA2011}.

Recent advancements in satellite sensing \citep{KN2017, LY2024, MY2022}, computer vision and multimodal technologies \citep{LW2022, MY2023}, and computational power \citep{HJ2018} have made it increasingly practical to obtain detailed geometric and indicative attributes for building instances (Table 1). This includes demand for multi-attributes such as rooftops \citep{GB2024, LZ2023, SQ2024, SW2021, ZZ2022, ZY2022}, height \citep{CQ2024, CY2024, LM2020, SX2024, WW2023}; function \citep{KB2024, ZY2024}, structure \citep{OY2023}, and age \citep{SM2022}. Despite the progress in multi-layered 3D city models at varying Levels of Detail (LoD) according to the CityGML international standard, large-scale 3D building data with geometric attributes remains scarce \citep{GY2023, TL2019, KT2009, GG2012}. Furthermore, the lack of standardized definitions and methods for extracting building indicative attributes has led to a predominant focus on geometric properties in current building data, hindering a thorough understanding of urban structures. Historically, building attributes like rooftops and heights were extracted at the grid scale, categorizing all buildings within a grid into one class, which proved inadequate for detailed urban analysis \citep{WW2023, LM2021}. Consequently, developing universal methods for the rapid extraction of building multi-attributes is vital for effective urban planning, research, and achieving Sustainable Development Goals (SDGs) 7, 9, 11, and 13.

\begin{table}[H]
\centering
\caption{Existing typical large-scale building datasets.}
\small
\resizebox{\textwidth}{!}{%
\begin{tabular}{|m{3cm}|m{4.5cm}|m{4.8cm}|m{2.8cm}|m{1.7cm}|m{3.8cm}|}
\hline
\textbf{Dataset} & \textbf{Source/Time Span} & \textbf{Coverage} & \textbf{Methods} & \textbf{Resolution} & \textbf{Type} \\
\hline
Microsoft BRA \citep{GB2024} & Bing map; No time span & Not include China & DNN & Vector & Rooftop \\
\hline
Google BRA \citep{SW2021} & Google map; No time span & \makecell[l]{Africa / South Asia and \\ Southeast Asia / Latin America} & U-net & Vector & Rooftop \\
\hline
CBRA \citep{LZ2023} & Sentinel 2; 2016-2021 & China & STSR-Seg & 2.5m & Rooftop \\
\hline
90-cities-BRA \citep{ZZ2022} & Google Earth satellite; 2020 & 90 cities in China & Deeplab-V3 & Vector & Rooftop \\
\hline
East Asia buildings \citep{SQ2024} & \makecell[l]{Google Earth satellite, \\ GUB2018; 2022} & \makecell[l]{China, Japan, South Korea, \\ North Korea and Mongolia} & CLSM & Vector & Rooftop \\
\hline
EUC \citep{LC2020a} & Landsat and Sentinel-1 SAR; 2015 & \makecell[l]{Europe, USA, China} & Machine learning & 1km² & Height \\
\hline
Wu et al., 2023 \citep{WW2023} & \makecell[l]{Sentinel 1-2, PALSAR, \\ LUOJIA1-01; 2020} & China & Machine learning & 10m & Height \\
\hline
Northern Hemisphere \citep{CY2024} & \makecell[l]{Sentinel-1/2 images; Google \\ Earth satellite; 2020} & \makecell[l]{China, the conterminous United \\ States (CONUS), Europe} & SRHS & 2.5m & Height \\
\hline
GABLE \citep{SX2024} & \makecell[l]{Beijing-3 satellite imagery, \\ WSF2019; 2023} & China & RPN & Vector & Rooftop and height \\
\hline
3D-GloBFP \citep{CY2024} & Microsoft BRA; 2020 & Global & Machine learning & Vector & Height \\
\hline
Zheng et al., 2024 \citep{ZY2024} & \makecell[l]{East Asian buildings, Baidu, \\ OSM, Gaode; No time span} & \makecell[l]{Three major urban \\ agglomerations in China} & Machine learning & Vector & Function \\
\hline
CBMA (Ours) & \makecell[l]{Google Earth satellite, \\ SpatialCites; 2021-2024} & China & OCRNet and XGBoost & Vector & \makecell[l]{Rooftop, height, function, \\ age and quality} \\
\hline
\end{tabular}
}
\end{table}

\FloatBarrier 

Despite significant efforts, extracting 3D building data remains fraught with challenges due to insufficient spatial and temporal resolution, limited training samples, and high costs. Currently, large-scale building footprints represented by Google and Microsoft \citep{GB2024, SW2021} are progressing towards building-instance-level detail. \citet{ZZ2022} and \citet{SX2024} produced rooftop data at the same level, but Zhang's work covers only 90 cities in China, and Sun's dataset, while nationwide, uses non-open-source data, limiting update ability. \citet{CQ2024} utilized open-source Google Earth imagery but did not provide imaging times for each region, and \citet{LZ2023} used super-resolution algorithms on Sentinel-2 data, complicating vectorization and direct application. High manual labeling costs restrict the number of high-resolution labels, limiting the coverage of diverse urban buildings \citep{BITC2021} and failing to reflect construction variations across different climatic zones \citep{SX2024}. Current methods for extracting building heights include LiDAR, SAR, and high-resolution optical remote sensing. LiDAR uses satellite, aerial, and vehicle-mounted lasers for high-resolution surveying but incurs high equipment costs \citep{PY2019}. SAR emits microwave signals and receives reflections, often suffering from mixed scattering effects and high data processing costs \citep{LM2020}. High-resolution optical remote sensing directly calculates building heights but lacks comprehensive 3D information due to the limitation of satellite angles and dates, making deep learning interpretations less interpretable \citep{LC2020a}. Combining building footprints with multi-source data, such as street view images (SVIs), shows promise but faces occlusion and incomplete coverage issues \citep{AJK2009, CH2021, CQ2021}. These methods highlight the need for continuing exploration of novel approaches to comprehensively capture the multi-dimensional attributes of urban structures. This is particularly crucial in developing countries where traditional aerial surveys are economically prohibitive, time-consuming, and involve high analytical costs. In contrast, open-source data presents greater potential for the rapid extraction of 3D building information \citep{CY2024, ZZ2022}.

Meanwhile, our current understanding of urban structures remains primarily at the physical level represented by existing building rooftops and height data due to the lack of comprehensive indicative attributes of buildings, such as function, age, quality, and vacancy. To the best of our knowledge, there are currently no large-scale datasets that provide these attributes at the building-instance level. Existing studies have primarily used SVIs data to estimate building function, age, and quality. However, due to the incomplete spatial distribution of SVIs data and the spatial mismatch between SVIs data and building data in previous studies \citep{LY2023}, the methods are difficult to scale up for national-scale data production. For building functions, it is limited to interpreting building functions only from remote sensing images. In previous studies, POIs data, building morphology data, and even Location Based Service (LBS) data have been used, but due to data coverage and the lack of large datasets, there are no national-scale data at present. Building age has greater application value in urban renewal studies. While the temporal resolution of remote sensing imagery represents the spatial distribution patterns of buildings within a specific time frame, building age requires precise information on the main construction year of each building. As for building quality, conventional remote sensing imagery cannot capture the characteristics of building quality. \citet{CJ2023} collected millions of SVIs from 2015 across China to measure street-level spatial disorder in over 700,000 streets across 264 cities. While this dataset includes some building quality features, it lacks specific building-level details and does not cover all street-facing or visible buildings due to sparse sampling POIsnts.

Accurate and comprehensive building data are essential for supporting digital and intelligent urban research and planning. The spatial distribution, three-dimensional information, function, quality, and construction age of buildings can reveal the fine-grained spatial and 3D dynamic evolution patterns of cities at both the physical and social functional levels \citep{DX2023}. This understanding is crucial for urban development, redevelopment, and the interaction between humans and the built environment \citep{YL2024}. By integrating geographic analysis packages, deep learning, and ensemble machine learning models, this paper proposes a geospatial intelligence framework for the rapid extraction of multi-attribute building information. Utilizing multi-source data, this study has produced, for the first time, a comprehensive multi-attribute building dataset (CMAB) covering 3,667 natural cities in China. The main contributions of this paper can be summarized as follows: (1) Enhanced accuracy of three-dimensional building products using multi-source data and ensemble machine learning; (2) The first national-scale dataset providing multi-attribute buildings at the instance level; (3) A methodological framework using open-source data for rapid acquisition of comprehensive building attributes; (4) Manually creation of extensive labeled data on building rooftops, heights, functions, ages, and quality, providing a foundation for further research and application.

\section{Methods}\label{secLR}

Our method involves four procedures: (1) preparation before prediction: we define the boundaries of spatial cities to define the extraction range of data products (see section \textbf{Study area and sampling strategy}). We select building samples based on climate zones and administrative city levels in China. All buildings are categorized into five classes according to their respective administrative levels. (2) extracting geometric attributes: rooftop samples, enhanced by manual labeling, are used to train the OCRNet model. A spatial aggregation method is employed to extract building rooftops for all spatial cities. Based on this, we calculate the morphological, density, and locational characteristics of buildings at different scales \citep{BFCY2022, CY2024}. Suitable models are trained for each administrative level class to obtain building heights, completing the extraction of 3D buildings (see section \textbf{Geometric attributes extraction}). (3) extracting indicative attributes: multi-source data is utilized to further extract functional characteristics at different scales, predicting building functions based on building height. By integrating impervious surface data from GAIA and sixty million SVIs, we assign building age and quality to each building instance through spatial matching and object detection (see section \textbf{Indicative attributes extraction}). (4) validate multiple attributes: we conducted the model evaluation of the building rooftop, height, and function with the validation dataset and also validated the building's height, function, quality, and age through manual SVIs labeling (see section \textbf{Technical validation}). The methodological framework of this research is shown in Figure 1, and details of each procedure will be explained below.

\begin{figure}[!htbp] 
\centering
\includegraphics[width=1\textwidth]{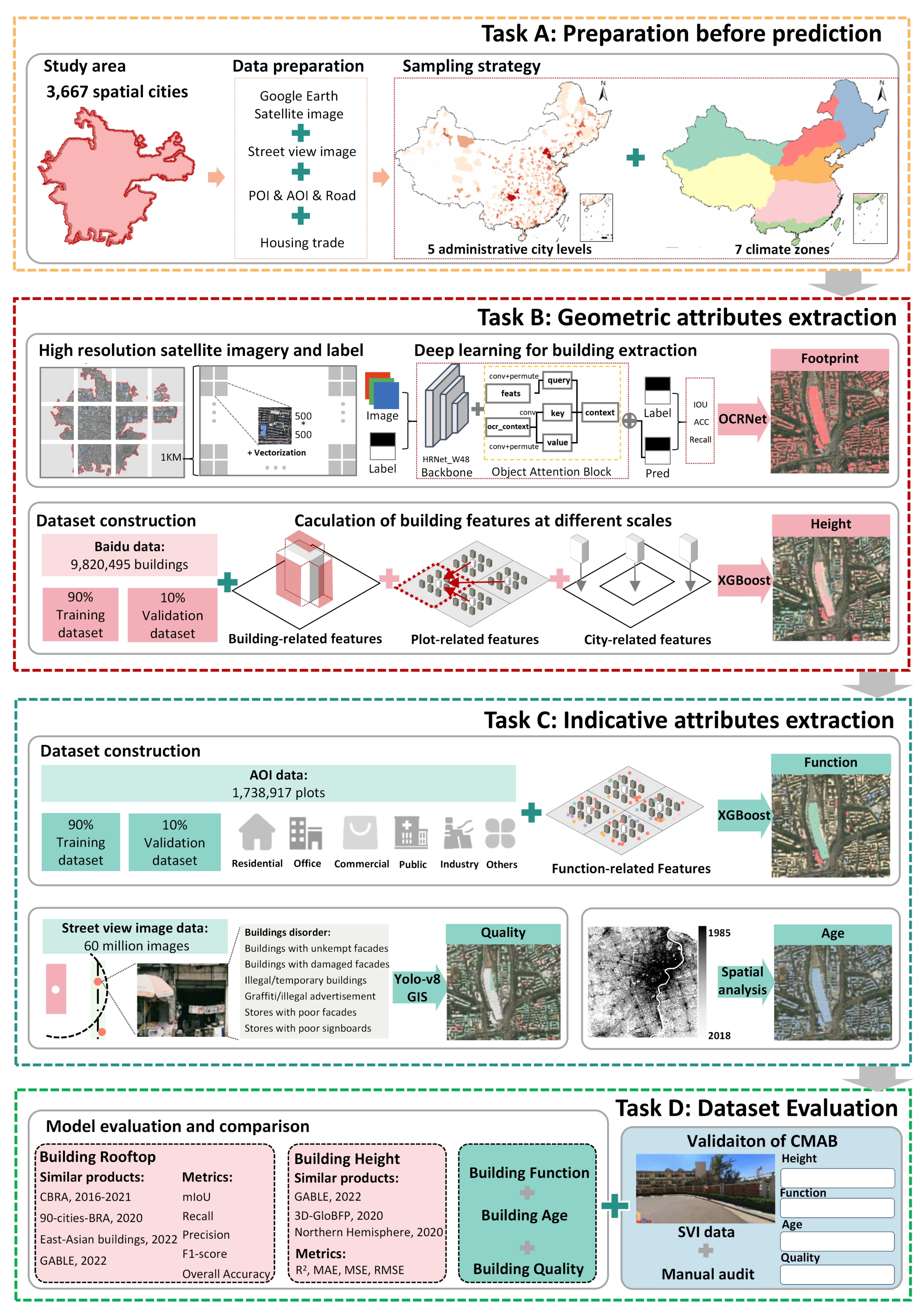} 
\caption{\textbf{The overall workflow of this study.}}
\label{fig:stream}
\end{figure}
\FloatBarrier 

\subsection{Study area and sampling strategy}
Training and inference across the entire land cover area of China are constrained by computational power and data limitations. The inclusion of large areas of non-human activity, which contain redundant information, can reduce model accuracy and increase the burden on computational resources. \citet{ZZ2022} improved computational efficiency by using built-up areas from the FROM-GLC30 data, while \citet{LZ2023} assumed that people primarily reside near basic administrative units and employed heuristic sampling using county-level administrative units in China. The administrative city boundaries used by Chinese government departments encompass extensive rural and non-built-up areas, making them unsuitable for spatial sampling strategies. To tackle these challenges, recent research has introduced the concepts of spatial city boundaries, functional urban areas, and Degree of Urbanization boundaries \citep{ML2019, ML2020, LQ2024}. From a building identification standPOIsnt, spatial city boundaries, derived from night light imagery, land use and land cover data, and related urban GIS datasets, offer greater accuracy and efficiency for urban studies. This paper follows the methodology of Ma and Long, utilizing the concept of spatial cities to define these boundaries across China \citep{ML2019}. Therefore, we utilized the boundaries of 3,667 spatial cities across mainland China, each with an area exceeding 2 km², to define building areas, encompassing a total area of 95,670 km² (with an average of 26.1 km² and a maximum of 5,121 km², 1\% of China's total land area of 9,634,057 km²) (Figure 2).

The completeness and bias of different data sources vary across regions (Figure 2). According to previous studies \citep{BC2023, PM2023, HB2023}, multi-source data such as AOI (Area of Interest), POIs (POIsnt of Interest), and SVI data are generally more complete in urban areas and more deficient in rural areas. Our statistics (Figure 2) indicate that the range of the multi-source data used aligns well with the spatial extent of the identified spatial cities. This congruence suggests that using spatial sampling based on urban entities helps mitigate the bias of open-source data across different regions, ensuring the comparability of building attribute predictions. To further improve the efficiency and accuracy of recognition, and considering the varying levels of investment and construction intensity in different cities, this study adopts the city classification system proposed by \citet{ZZ2022}. We classify the administrative locations of all identified building roof centroids into five categories based on China's urban administrative hierarchy: (1) 6 municipalities directly under the central government and special administrative regions, (2) 28 provincial capitals and five planned cities, (3) 261 prefecture-level cities, (4) 388 county-level cities, and (5) Non-urban areas (buildings not located within any administrative city). The height and function models are trained based on these administrative divisions.

\begin{figure}[h!] 
\centering
\includegraphics[width=1\textwidth]{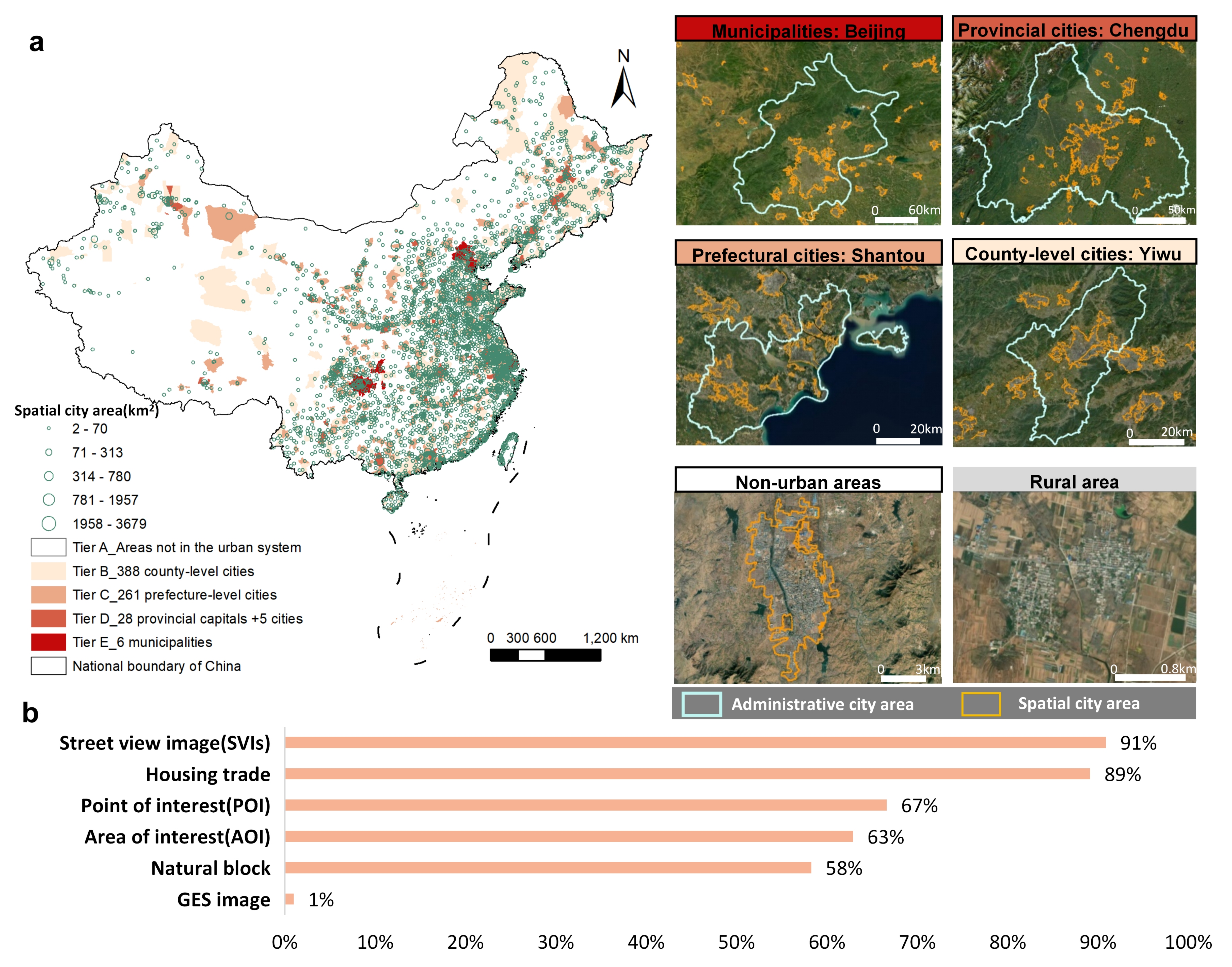}  
\caption{\textbf{Sampling with spatial cities in China and the completeness of multi-source data.} \small{(a) Spatial city and administrative city distribution. (b) The ratio of multi-source data within physical cities to the total multi-source data across China. The diagram shows that most of the data in different datasets are concentrated in spatial cities that only occupy 1\% area of China.}}
\label{fig:stream}
\end{figure}
\FloatBarrier 

\subsection{Data source}
\subsubsection{Google Earth Satellite image}
We collected high-resolution remote sensing imagery from Google Earth Satellite (GES) at a resolution of 0.3 m per pixel, with original resolutions under 1 m in remote areas. These images were downloaded in March 2024 using scripts from the Google Earth API. Since GES imagery comes from various global remote sensing satellites (WorldView, QuickBird, and SPOT), there are regional differences. Considering the high temporal and spatial costs of acquiring long-term GES imagery, for each spatial city, we selected the most recent and cloud-free images from the past five years through manual visual comparison. Based on the timestamps of the centroid POIsnts of each spatial city, we obtained the temporal distribution of all city images, with 70\% of the images taken between 2022 and 2024 (Figure S4).

\subsubsection{Other source datasets}
Existing research indicates that building attributes can be obtained from various data sources such as remote sensing imagery, street view imagery, housing statistics, and urban geographic data \citep{BFCY2022, LQ2024, AY2017, CQ2021, CW2023b}. This study employs and extends the widely used indicator systems from previous research, utilizing data from different sources and volumes (Table 2). Considering data availability, we have summarized and visualized the primary data sources and validation data sources used in this study (Figure 2).

\begin{table}[ht]
\centering
\caption{Multi-source datasets used in the workflow.}
\small
\resizebox{\textwidth}{!}{%
\begin{tabular}{|m{4cm}|m{2cm}|m{2cm}|m{4cm}|m{3cm}|m{6.5cm}|}
\hline
\textbf{Datasets} & \textbf{Time} & \textbf{Type} & \textbf{Resolution} & \textbf{Number/Storage} & \textbf{Source} \\
\hline
GES Imagery & 2023 & Raster & 0.3-1m/500*500 pixels & 5E8 / 20TB & \makecell[l]{Google Earth (https://www.google.cn/)} \\
\hline
Streetview Image & 2014-2023 & Raster & 800*400 pixels & 6E7 / 14TB & \makecell[l]{Baidu Maps (https://map.baidu.com/)} \\
\hline
Building Height & 2023 & Polygon & Vector & 1.7E7 / 8GB & \makecell[l]{Baidu Maps (https://map.baidu.com/)} \\
\hline
POIs & 2021 & POIsnt & Vector & 1.4E7 / 3GB & \makecell[l]{Baidu Maps (https://map.baidu.com/)} \\
\hline
AOI & 2023 & Polygon & Vector & 8.3E6 / 7GB & \makecell[l]{Baidu Maps (https://map.baidu.com/)} \\
\hline
Natural Cities & 2020 & Polygon & Vector & 3.7E3 / 108MB & \makecell[l]{BCL (https://www.beijingcitylab.com/)} \\
\hline
Adminstrative Boundaries & 2020 & Polygon & Vector & 6.9E2 / 17MB & \makecell[l]{Earth data(www.natural.earthdata.com/)} \\
\hline
Climte Zone & 2017 & Polygon & Vector & 7 / 6MB & \makecell[l]{Yao et al., 2018} \\
\hline
Rooftop Labels & 2021 & Raster & Vector & 1.5E4 / 5GB & \makecell[l]{Fang et al., 2021} \\
\hline
Road & 2019 & Polyline & Vector & 1.3E7 / 2GB & \makecell[l]{Gaode (https://lbs.amap.com)} \\
\hline
Natural Blocks & 2020 & Polygon & Vector & 1.0E6 / 637MB & \makecell[l]{Geohey (https://geohey.com)} \\
\hline
Housing Trade & 2023 & Text & POIsnt & 3.8E6 / 855MB & \makecell[l]{Anjuke (https://www.anjuke.com)} \\
\hline
\end{tabular}
}
\end{table}

\begin{figure}[!htbp] 
\centering
\includegraphics[width=1\textwidth]{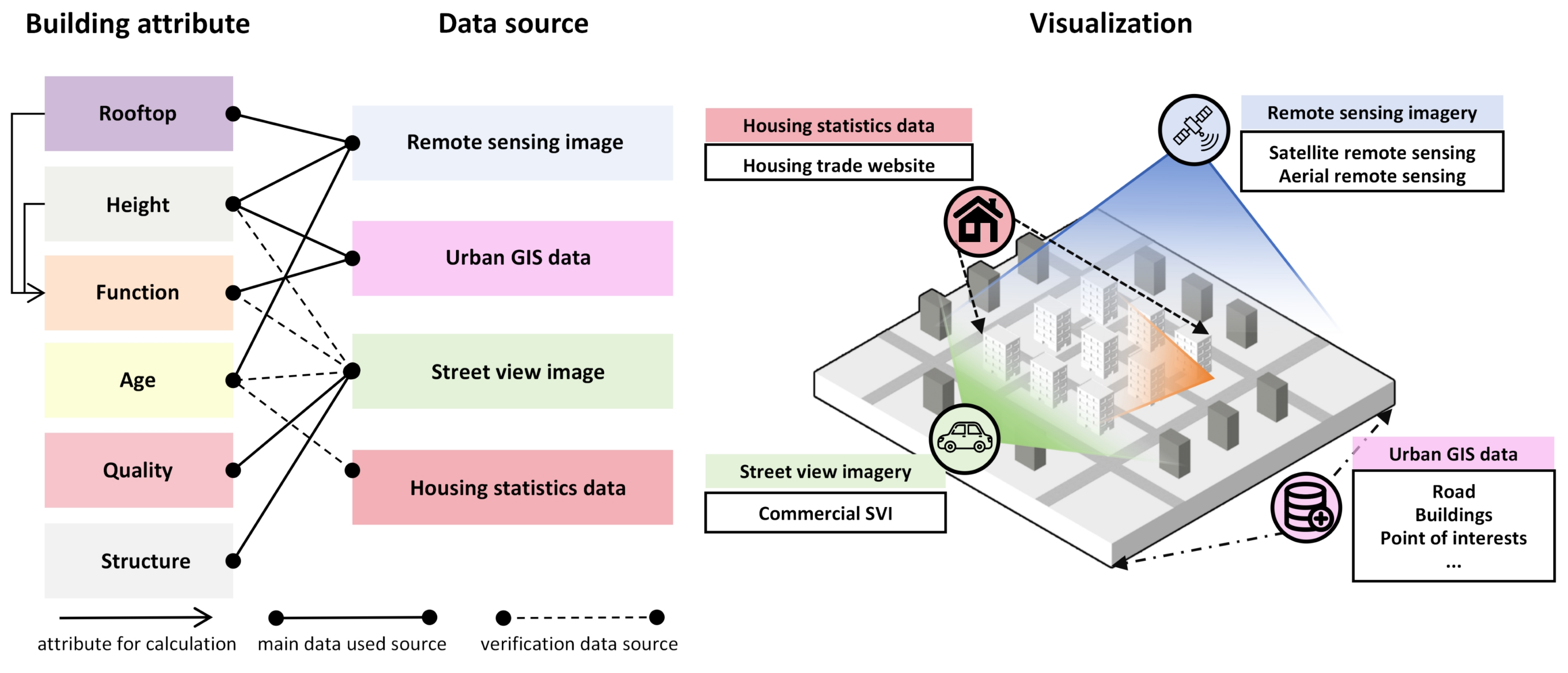}  
\caption{\textbf{Building attributes and visualized source. }\small{"building attribute" records the attributes calculated, and the "data source" is the type of data source visualized for calculation.}}
\label{fig:stream}
\end{figure}
\FloatBarrier 

\subsection{Building partition model and combination model}
This study utilized building data from 85 cities in Baidu's 2023 dataset as ground truth for constructing a machine learning model to predict building heights based on multi-scale building features. This model generated three-dimensional building data for the entire country, and the accuracy of this data was subsequently evaluated. Upon generating the three-dimensional building data for Chinese cities, additional features related to building functions and three-dimensional morphological attributes were extracted from the height data. Using these comprehensive building features and treating the 2023 Baidu AOI data's functional categories as ground truth, a secondary machine learning model was developed to predict building functions (see Supplementary Table 1, 2). This model produced functional attribute data for three-dimensional buildings nationwide, and the accuracy of this data was assessed.

As outlined, we predicted two attributes for each building roof: building height (meters) and building function. Employing parallel processing, distributed computing, and hardware optimization (such as GPU acceleration), the XGBoost algorithm \citep{CG2016} demonstrated superior training efficiency over traditional tree models (e.g., random forests) when handling large-scale and high-dimensional data. Consequently, the GPU-accelerated XGBoost algorithm (Python library XGBoost 2.0.3) using NVIDIA GeForce RTX 3070 (Python library cupy 13.1) was employed for model training and prediction on this extensive building dataset, comprising tens of millions of training samples and hundreds of dimensional features.

Ensemble learning enhances model accuracy and stability by amalgamating the predictions of multiple models \citep{BR1996, FJ2000}. It encompasses three primary methods: Bagging, Boosting, and Stacking. The approach we utilized can be described as 'Bootstrap Aggregated XGBoost,' which combines aspects of Bagging and Boosting. Multiple training subsets were generated through bootstrap sampling, with each subset used to train an independent XGBoost model. The predictions from these models were then averaged (or voted upon) to enhance overall performance. To mitigate the issue of uneven height and feature sample distributions, bootstrap sampling with replacement was employed \citep{EB1993}. This technique facilitated the creation of 100 models based on various data partitioning methods and XGBoost parameters for each height partition model and combined model, categorized according to city administrative levels. This methodology aimed to enhance the accuracy of overall model estimates and elucidate the uncertainty in model predictions.

\subsection{Geometric attributes extraction}
\subsubsection{Building rooftop}
Regarding the building rooftop, existing building instance segmentation labels often rely on segmentation models for assistance, resulting in lower accuracy in some regions \citep{BITC2021}, and the remote sensing images used are difficult to obtain and lack complete coverage \citep{HX2022}. Moreover, high-resolution image labels are rarely open-sourced \citep{ZZ2022}. In this study, we selected the BITC dataset as the manually labeled data for building rooftop segmentation. This dataset comprises 7,260 slices (500 pixels * 500 pixels) with 63,886 buildings, labeled from 0.3 m resolution Google high-resolution remote sensing images, covering Beijing, Shanghai, Shenzhen, and Wuhan.

We supplemented the BITC building rooftop labels with additional manual labeling based on China's climatic zones, as BITC only focuses on four cities and does not represent the entire architectural styles of China. It also pays less attention to densely built urban areas such as urban villages and other high-density urban areas. China's architectural types are influenced by diverse natural environments and climates, resulting in varied cultural styles and complex structural layouts (Figure S2). Additionally, current satellite images exhibit blurred visual features \citep{CD2022}. Based on the BITC data, we used Baidu building datasets extraction annotation areas and selected appropriate urban dense areas for labeling according to China's building climate zoning standards, ensuring an equal number of slices in each zone to maintain sample balance (Figure 4). Finally, we used 8,760 slices as the complete annotated dataset, including 6,973 training slices and 1,787 validation slices, with a total of 114,783 buildings distributed across seven cities (Beijing, Urumqi, Hefei, Shenyang, Hohhot, Lhasa, and Xiamen) and three BITC original cities (Shanghai, Shenzhen, and Wuhan). The final data augmentation operations included random cropping, image rotation, color jittering, image blurring, and noise addition. The dataset consists of annotation files in MS COCO 2017 format and corresponding binary building mask images, providing foundational data for high-resolution remote sensing image research on building detection and extraction (Figure S2).

We used OCRNet to extract building rooftops from standardized and preprocessed GES images. Unlike the Deeplabv3p method employed by \citet{ZZ2022}, which focuses on the relationships between context pixels without explicitly utilizing features of the target area, the OCRNet method addresses the problem of object region classification rather than pixel classification. That is, the OCRNet method explicitly enhances object information. Therefore, OCRNet is superior in terms of performance and complexity (see Supplementary note 1).

During the inference stage, each city's remote sensing image is divided into grid slices. These slices are then sequentially input into the model for segmentation, yielding segmentation results for each grid. The Douglas-Peucker algorithm is applied to vectorize the raster data. Finally, the segmentation results of all grids are spatially merged. Because the same building may be divided because it is located in different slices, we use a post-processing method to eliminate these cracks. The specific method is to detect the similarity of polygon edges of buildings in the corresponding buffer zone of the boundary and repair them with fishing nets.

\subsubsection{Building height}
Related studies have verified that Baidu data meets the accuracy requirements for building height modeling in China \citep{LM2021, SX2024, WW2023}. According to \citet{LM2021}, the overall mean height deviation of Baidu building data is 1.02 meters, with an accuracy rate of 86.78\%. In this study, we obtained building data for 96 major cities in China from the Baidu Map service (\url{www.map.baidu.com/}), which includes 12,772,156 individual building instances with floor information. After data cleaning, 9,820,495 buildings remained, of which 10\% (982,049 buildings) were used for validation, and 8,838,446 buildings were included in the training dataset (Figure S2).

Existing research indicates that building height is correlated with the morphological patterns of building rooftops, the state of neighboring buildings, adjacent streets, and the morphology of the associated blocks \citep{CY2024, BL2017, LZ2023, TA2024}. For example, the morphological pattern of building footprints affects the complexity of building heights, with taller buildings typically having larger base areas, while shorter buildings tend to have more neighbors \citep{BL2017}. Buildings adjacent to streets and main roads may have greater heights due to skyline control and commercial development \citep{TA2024}. High-density (or compact) streets may imply more high-rise buildings to accommodate a larger potential population; buildings within the same block are more likely to have similar heights \citep{LY2023}. Additionally, we posit that building height is also related to the location of the area and the intensity of investment and construction \citep{ZZ2022}. Therefore, we have included the relationship between buildings and streets, their location within different administrative scales, and their relationship to urban functional centers (see Supplementary note 2 and Figure S4).

The quantification of model uncertainty is essential for interpreting results, with primary sources being the training data and model parameters. \citet{LM2020} employed an ensemble of multiple trees in a random forest to mitigate bias and overfitting inherent in single models. They recorded the standard deviation and mean of 100 predictions per sample, thus deriving the coefficient of variation as a measure of prediction uncertainty. In contrast, XGBoost does not provide independent prediction results from multiple trees. Hence, we randomly selected 10\% of the test data to assess error and uncertainty (see Figure S5).

\subsection{Indicative attributes extraction}
\subsubsection{Building function}
This study constructs a training set of building function data based on the identification of building heights. According to existing research \citep{DY2022, ZY2024}, predicting building functions typically involves combining building morphology with other data sources. Previous studies have often relied on datasets derived from manual function labels or building information provided by OSM maps. However, OSM data in China suffers from limited coverage and accuracy, and manually labeled data is challenging to scale nationwide. Fortunately, multi-source open big data, such as Area of Interest (AOI) data, provide plot-level functional characteristics. In China, buildings of different functions are often clustered by plots. Thus, this study determines building functions using Baidu's 2023 AOI data, which offers functional features for 30 categories of plots (see Supplementary Table 1). The functions are reclassified, assuming that all buildings within a plot share the same function as the plot itself (see Supplementary Table 2).

\begin{figure}[!htbp] 
\centering
\includegraphics[width=1\textwidth]{figures/figures_04.pdf}  
\caption{\textbf{Construction of building features for height and function estimation.} \small{The features marked in bold are categorical and are encoded using label encoding and the category type to enable XGBoost >1.3 to automatically recognize categorical features. The features marked in red indicate new characteristics in the function index system compared to the height index system. See \textbf{Supplementary Table 2} for more details.}}
\label{fig:stream}
\end{figure}
\FloatBarrier 

Studies have demonstrated that building functions can also be inferred from morphological features in multi-source data such as POIs, social media, and SVIs. On the one hand, buildings with different functions exhibit distinct morphological characteristics. For instance, public and commercial complex buildings typically have larger volumes, while office buildings tend to have high floors. Additionally, similar buildings often cluster together in China, allowing the characteristics of the block on which a building is located to help infer its function. The locational attributes of buildings and the distribution characteristics of surrounding POIss can also partially predict building functions. Notably, previous studies have utilized SVIs data and social media data to infer building functions. However, the former only covers buildings along streets, and the latter is challenging to apply comprehensively to entire urban areas, limiting their generalizability. Therefore, this study ultimately constructed a predictive variable system from four dimensions: building morphological characteristics, block characteristics, urban locations, and the distribution features of 19 types of POIss such as life services and transportation services. The final set of features comprises 91 variables, some of which overlap with those used to predict building height (Figure 4).

\subsubsection{Building quality}
Our study builds on the existing work by extending Chen's \citep{CD2022} methodology with the enhancements introduced by \cite{LY2024}. We applied the updated Yolov8 deep learning model to analyze six specific indicators related to building quality: “Buildings with unkempt facades,” “Buildings with damaged facades,” “Illegal/temporary buildings,” “Graffiti/illegal advertisement,” “Stores with poor facades,” and “Stores with poor signboards.” \citet{CD2022} originally collected 4,876,952 SVIs from 264 cities in China, covering 1,219,238 sampling POIsnts across 70 million streets, demonstrating the potential for large-scale, human-eye scale assessments of street spatial quality. \citet{LY2024} further refined this approach by adding self-collected SVIs data and improving recognition accuracy through the updated Yolo-v8 model. To comprehensively represent the street-facing buildings in Chinese cities, we obtained all Baidu SVIs from 2014 to 2023, totaling 11,286,209 sampling POIsnts and 60 million images (14TB), covering 3,224 spatial cities. The quality of each building instance over the past decade was assessed using Yolo-V8 with a parameter setting of 0.25. Due to inconsistent spatial coverage of SVIs across different years, we used the most recent year with SVIs available within each building's buffer zone as the final quality assessment result (accelerated computation using Python library Vaex 4.17).
 The building disorder level for each building $i$ is represented by the total score of the average values of the relevant disorder categories for all street viewPOIsnts within a 100m buffer of the building $i$ centroid for each year $y$. The building quality $Q_{iy}$ \ref{eq:Tyk},\ref{eq:Qiy} is the total building disorder score of the building $i$ in year $y$ (Figure 5).
\begin{align}
T_{yk} &= \sum_{m=1}^{M} \left( \frac{\sum_{n=1}^{N_{ym}} S_{yknm}}{N_m} \right) \label{eq:Tyk} \\
Q_{iy} &= \sum_{k=1}^{6} \frac{T_{yk}}{M} \label{eq:Qiy}
\end{align}

$M$ means the total number of street viewPOIsnt in the buffer zone of each building centroid;  means the total number of street view images in the street viewPOIsnt $m$ in year $N_{ym}$; $k$ means the type of building disorder; $S_{yknm}$ means the score (0 or 1, exist or not exist) of the building disorder type $k$ of the street view image in the street viewPOIsnt $m$ in year $y$; $T_{yk}$ \ref{eq:Tyk} means the total disorder score of building disorder type $k$ within all street view POIsnt $m$ in the buffer zone of the building $i$ in year $y$; 
\begin{figure}[!htbp] 
\centering
\includegraphics[width=1\textwidth]{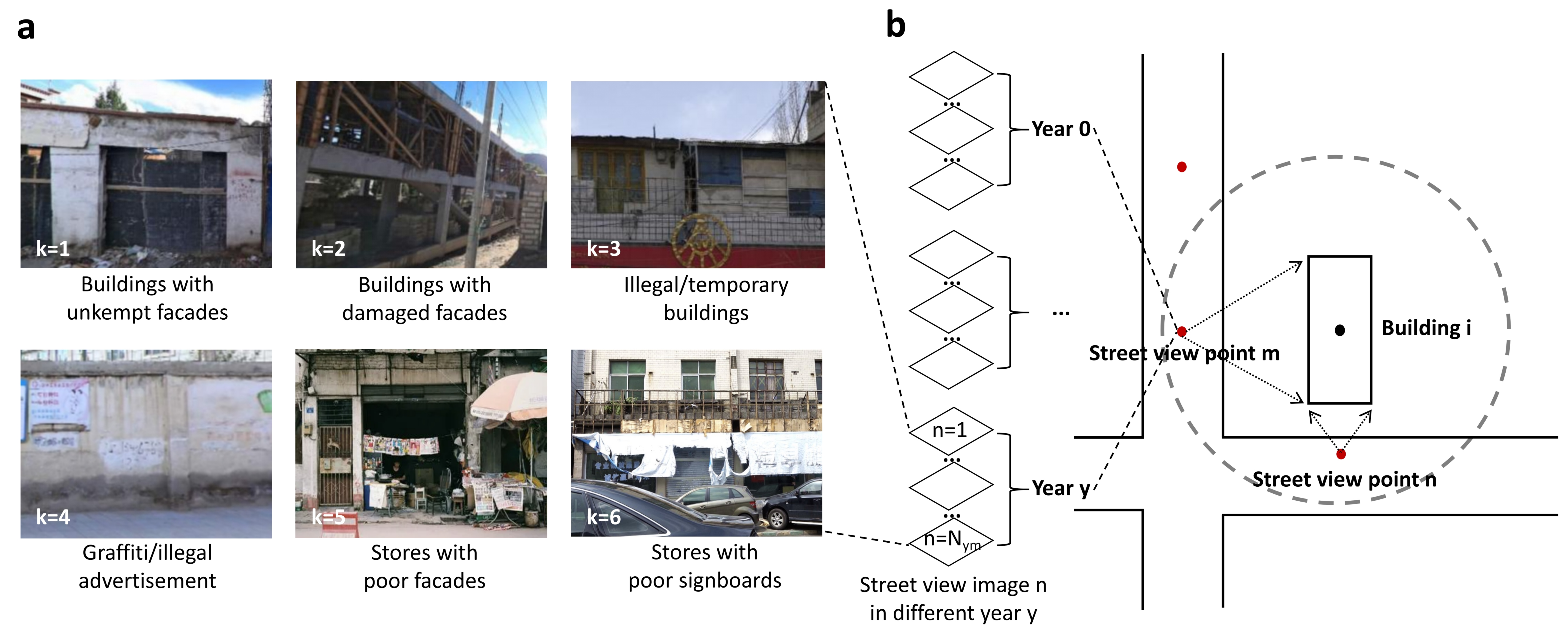}  
\caption{\textbf{Evaluating the quality of buildings along the street through SVIs.} \small{(a) Building disorder types k for building quality. (b) Temporal and spatial distribution of SVI in the building buffer zone. }}
\label{fig:stream}
\end{figure}
\FloatBarrier 

\subsubsection{Building age}
Existing studies have identified the age of street-facing buildings using SVIs \citep{SM2022, OY2023}. However, considering that such methods are challenging to scale up to a national level (as they require the age of all buildings rather than just street-facing ones), we employed long-term impervious surface data to determine the age of each building instance. Impervious surfaces consist of human-made structures that impede the natural infiltration of water into the soil, including rooftops, pavements, roads, etc. By reviewing existing research on impervious surface data and built-up area data \citep{GP2020, LX2020b}, we selected the GAIA data (1985-2018) with relatively high spatiotemporal resolution to determine the construction age of each building instance. We assumed that the expansion of impervious surfaces is synchronous with the construction age of buildings. Thus, by identifying the first appearance of a building instance's centroid in the spatial location of the impervious surface, we can assign 35 age categories to that building instance (Figure 6).
\begin{figure}[!htbp] 
\centering
\includegraphics[width=1\textwidth]{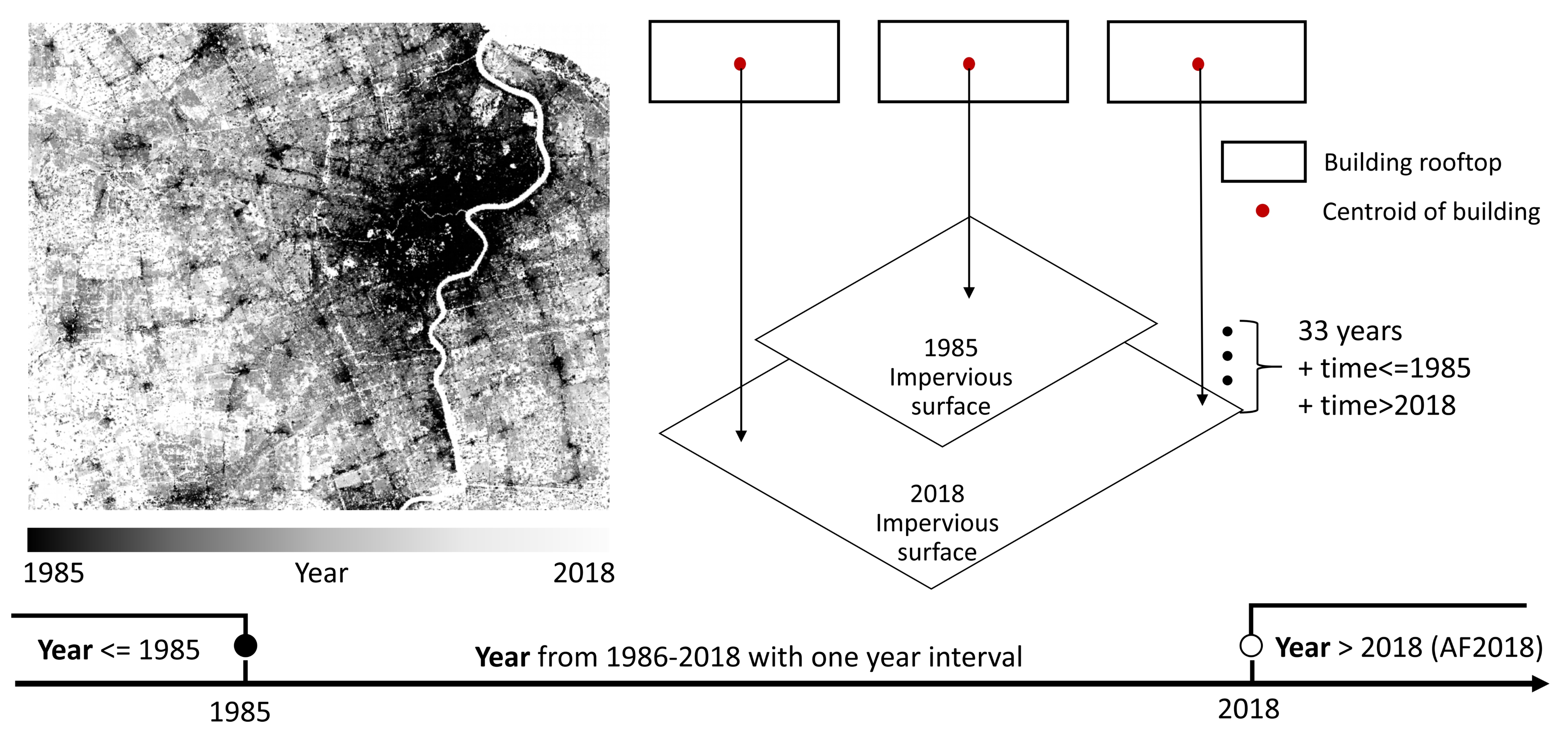}  
\caption{\textbf{Method for adding building age to buildings with GAIA data.}}
\label{fig:stream}
\end{figure}
\FloatBarrier 

\section{Data Records}
Overall, the data product we produced contains 29 million buildings covering the urban area of China, with a total storage space of 272GB. The product is organized by provinces and natural cities and saved in a standard GIS format. Each building rooftop is preserved as a polygon drawn by a limited number of POIsnts in a geographic coordinate system (WGS1984), including the building rooftop, height, function, age, and quality as building attributes, as shown in Figure 7. See \textbf{Supplementary Table 3} for the description of the attributes’ fields in the data. The entire data product can be publicly obtained through the Mendeley website after the publication of the paper.

\begin{figure}[!htbp] 
\centering
\includegraphics[width=1\textwidth]{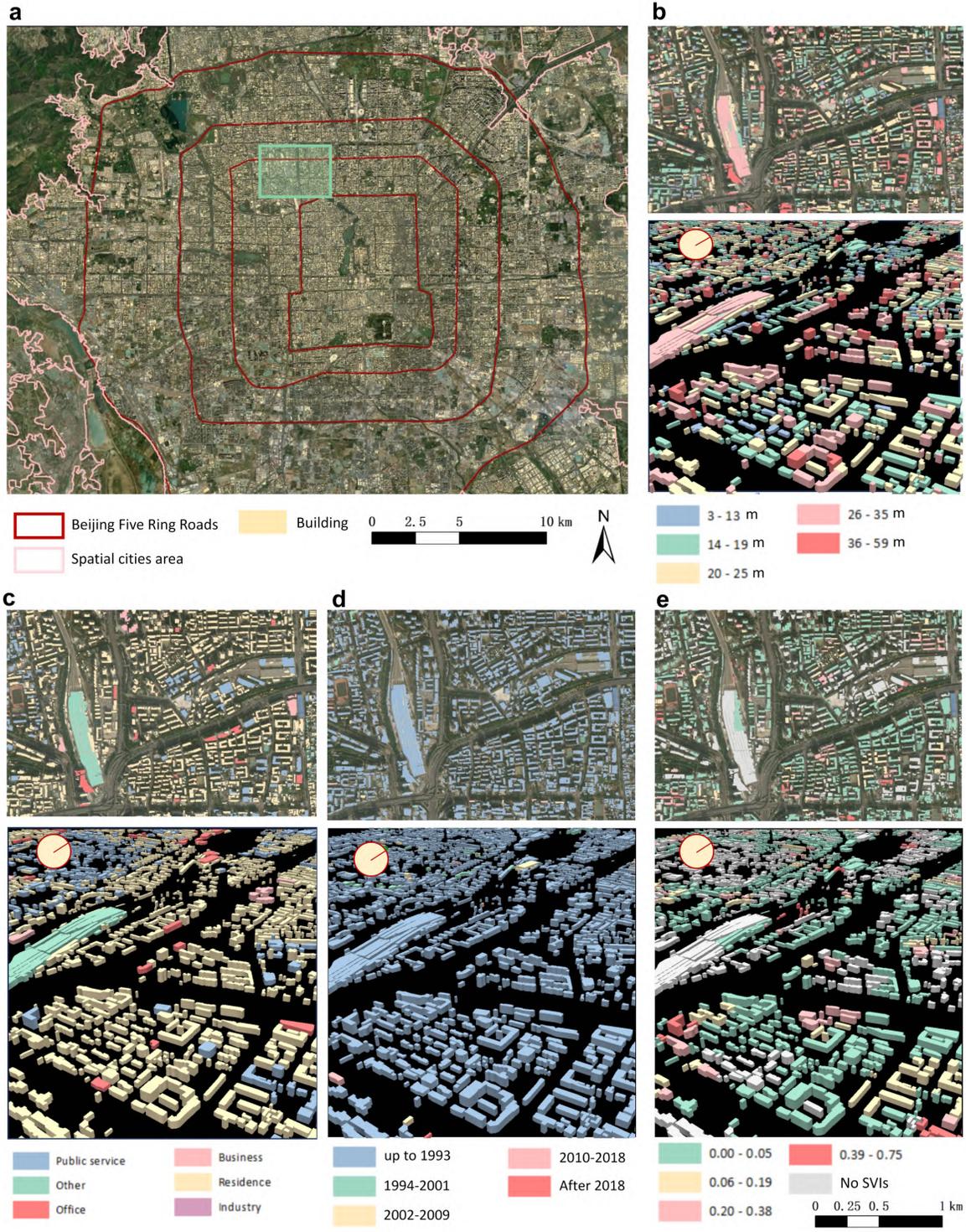}  
\caption{\textbf{Visualization of multi-attribute building dataset in Beijing sample.} \small{(a) sample regional distribution. (b) Building height visualization. (c) Building function visualization. (d) Building age visualization. (e) Building quality visualization.}}
\label{fig:stream}
\end{figure}
\FloatBarrier 

\section{Technical Validation}
The technical validation of the CMAB dataset is conducted in three parts: (1) the performance of the OCRNet model and the XGBoost model on the test set (including rooftop, height, and function); (2) comparison of our data with related published datasets (including rooftop, height, and functions); (3) validation by comparison of the predicted values with values directly observed from SVIs (including height, function, and age). For details, see the sections "Model evaluation and comparison for geometric attributes," and "Model evaluation and comparison for Indicative attributes."

For evaluation metrics of building rooftops, mIoU (mean Intersection over Union) represents the average segmentation accuracy across all classes. Accuracy denotes the overall pixel classification accuracy. The F1-score combines precision and recall, making it particularly suitable for imbalanced datasets. Precision and Recall provide the segmentation performance for each specific class, helping to identify which classes the model performs better or worse on. We use Precision \ref{eq:precision}, Recall \ref{eq:recall}, F1-score \ref{eq:f1score}, and Accuracy \ref{eq:accuracy} to evaluate the rooftop segmentation model:

\begin{align}
\text{Precision} &= \frac{TP}{TP + FP} \label{eq:precision} \\
\text{Recall} &= \frac{TP}{TP + FN} \label{eq:recall} \\
\text{F1-score} &= \frac{2 \times \text{Precision} \times \text{Recall}}{\text{Precision} + \text{Recall}} \label{eq:f1score} \\
\text{Accuracy} &= \frac{TP + TN}{TP + TN + FP + FN} \label{eq:accuracy} 
\end{align}

TP is the True Positives for class $i$ (building and background), FP is the False Positives for class $i$, and FN is the False Negatives for class $i$. mIoU \ref{eq:mIoU} is the mean IoU \ref{eq:IoU} of the building and background classes, with $k=2$ being the number of classes. Model accuracy metrics (RMSE/MSE/MAE/$R^2$) were evaluated on the building height:
\begin{align}
\text{mIoU} &= \frac{1}{k} \sum_{i=1}^{k} \text{IoU}_i \label{eq:mIoU} \\
\text{IoU}_i &= \frac{1}{k} \sum_{i=1}^{k} \frac{TP_i}{TP_i + FP_i + FN_i} \label{eq:IoU} 
\end{align}
For evaluation metrics of building height, model accuracy metrics (RMSE/MAE/R²) were evaluated on the building height. RMSE \ref{eq:rmse} emphasizes large errors and their impact. MAE \ref{eq:mae} reflects overall accuracy by averaging absolute errors. R² \ref{eq:r2} shows how well predictions fit the actual data. The formulas are as follows:
\begin{equation}
\text{RMSE} = \sqrt{\frac{1}{n} \sum_{i=1}^{n} (y_i - \hat{y}_i)^2}
\label{eq:rmse}
\end{equation}
\begin{equation}
\text{MAE} = \frac{1}{n} \sum_{i=1}^{n} |y_i - \hat{y}_i|
\label{eq:mae}
\end{equation}
\begin{equation}
R^2 = 1 - \frac{\sum_{i=1}^{n} (y_i - \hat{y}_i)^2}{\sum_{i=1}^{n} (y_i - \bar{y})^2}
\label{eq:r2}
\end{equation}

For the uncertainty of building height estimation, we randomly selected 10\% of the test data to assess error and uncertainty. For the remaining 90\% of the data, 20\% was randomly chosen as the validation set and 80\% as the training set in each iteration. This process was repeated 100 times, with XGBoost hyperparameters optimized through grid search during each iteration. The mean of 100 prediction results per building served as the final height prediction. Model accuracy metrics (RMSE/MSE/MAE/$R^2$) were evaluated on the test set. Uncertainty was quantified as the range of relative error $RE_i$ of the building $i$, each trained on different data splits and optimized hyperparameters. A wide range indicates high uncertainty, while a narrow range suggests consistent predictions.

Specifically, for each building sample $i$, $RE_i$ \ref{eq:RE} is defined as the ratio of the difference between the true building height $T_i$ and the predicted value $P_{i,j}$. Additionally, we provide the absolute error $AE_i$ and the range of absolute errors across 100 model estimates. Here, $j$ denotes the $j$ th predicted value of the 100 model estimation:
\begin{align}
\text{RE}_i &= \left| \frac{P_{ij} - T_i}{T_i} \right| = \left| \frac{{AE}_i}{T_i} \right| \label{eq:RE}
\end{align}

In order to verify the model reasoning and calculation results of each attribute, the study utilized SVIs to further verify the height, function, and age of buildings along the streets. Initially, five administrative cities were selected, and buildings along the streets were sampled for validation. These cities encompass different urban hierarchies, provinces, and climate zones. The sampling aimed to cover a diverse range of building heights, functions, and ages. Subsequently, the nearest POIsnt on the building's outline to the closest SVI POIsnt was designated as the observation POIsnt. The direction of the street view sampling was defined as vector 1, and the direction from the street view POIsnt to the observation POIsnt was defined as vector 2. The angle difference between these two vectors was calculated: if it fell between 45 and 135 degrees, the right-side image in the forward direction of the street viewPOIsnt was extracted; if it fell between -45 and -135 degrees, the left-side image was extracted. Finally, a manual auditing platform was established, involving an auditor with an urban planning background (Figure 8). This process resulted in the manual annotation of 2500 pieces of information regarding building height, function, and age.

\begin{figure}[!htbp] 
\centering
\includegraphics[width=1\textwidth]{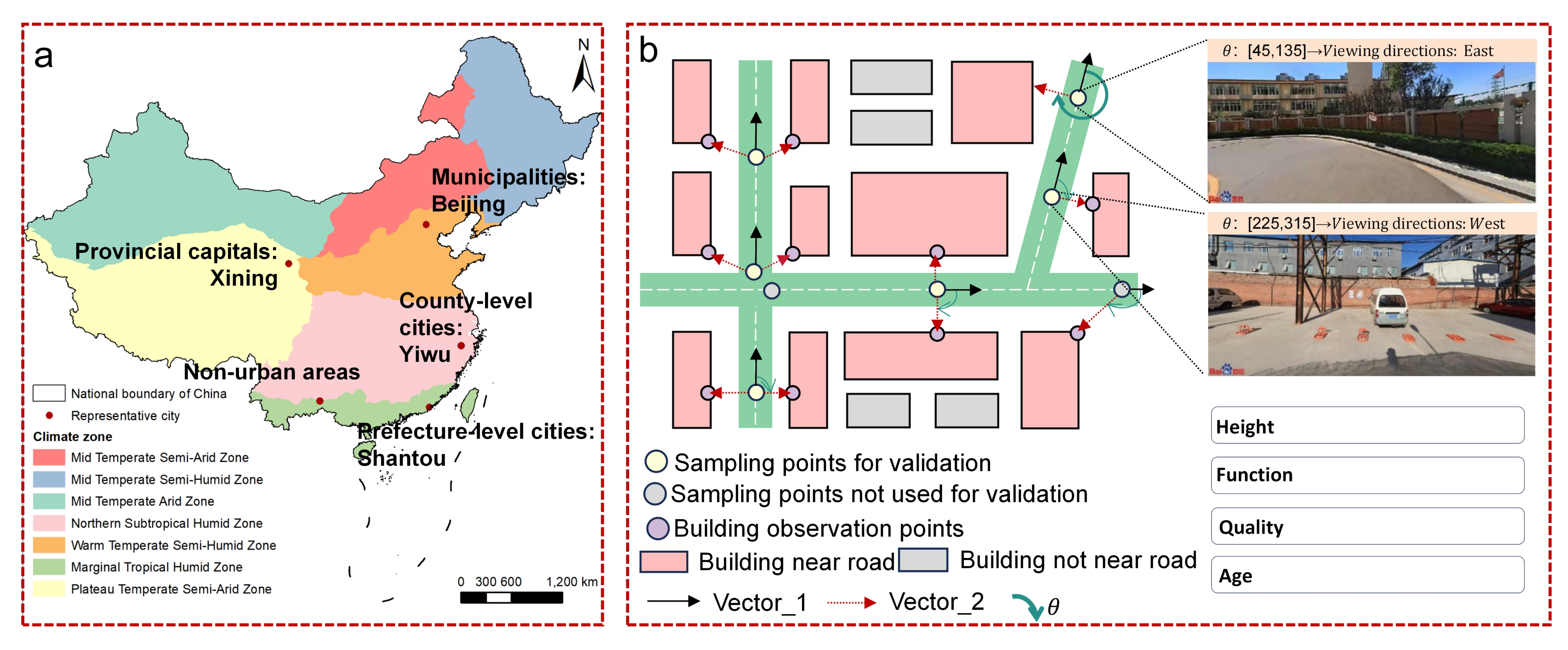}  
\caption{\textbf{Validation method of multi-attribute buildings.} \small{(a) Five representative cities distribution. (b) The matching process between street view images and buildings.}}
\label{fig:stream}
\end{figure}
\FloatBarrier 

\subsection{Model evaluation and comparison for geometric attributes}
\subsubsection{Building rooftop}
To evaluate the accuracy of our product, we utilized higher-resolution remote sensing imagery and supplemented the annotated dataset with dense urban areas. We based our evaluation on a validation set comprising 23,415 manually labeled building roofs from seven cities located in different climatic zones. The results demonstrate that our building roof segmentation model outperforms existing studies in terms of mIoU, Recall, and Precision on related datasets. After supplementing the annotated data to include 114,783 building instances, our rooftop segmentation model achieved an Accuracy of 91.59\%, a mIoU of 81.95\%, an F1 score of 89.93\%, and a Kappa coefficient of 79.86\% (Table 3), proving the model's capability to accurately identify buildings across various regions in China.

\begin{table}[H]
\centering
\caption{Data comparison from the evaluation metrics of rooftop extraction results.}
\small
\resizebox{\textwidth}{!}{%
\begin{tabular}{|m{5cm}|m{1.5cm}|m{1.5cm}|m{1.5cm}|m{1.5cm}|m{1.5cm}|}
\hline
\textbf{Dataset} & \textbf{mIoU} & \textbf{Recall} & \textbf{Precision} & \textbf{F1-score} & \textbf{Accuracy} \\
\hline
CBRA (Liu et al., 2023) & 0.63 & 0.79 & -- & 0.74 & 0.83 \\
\hline
90-city BRA (Zhang et al., 2022) & 0.58 & 0.79 & 0.88 & 0.83 & -- \\
\hline
GABLE (Sun et al., 2024) & 0.76 & -- & 0.75 & 0.81 & 0.88 \\
\hline
East Asian buildings (Shi et al., 2024) & -- & 0.85 & 0.81 & 0.82 & 0.87 \\
\hline
Ours (Without label enhancement) & 0.79 & 0.86 & 0.87 & 0.88 & 0.89 \\
\hline
Ours (With label enhancement) & 0.82 & 0.91 & 0.91 & 0.90 & 0.92 \\
\hline
\end{tabular}
}
\end{table}

\begin{figure}[!htbp] 
\centering
\includegraphics[width=1\textwidth]{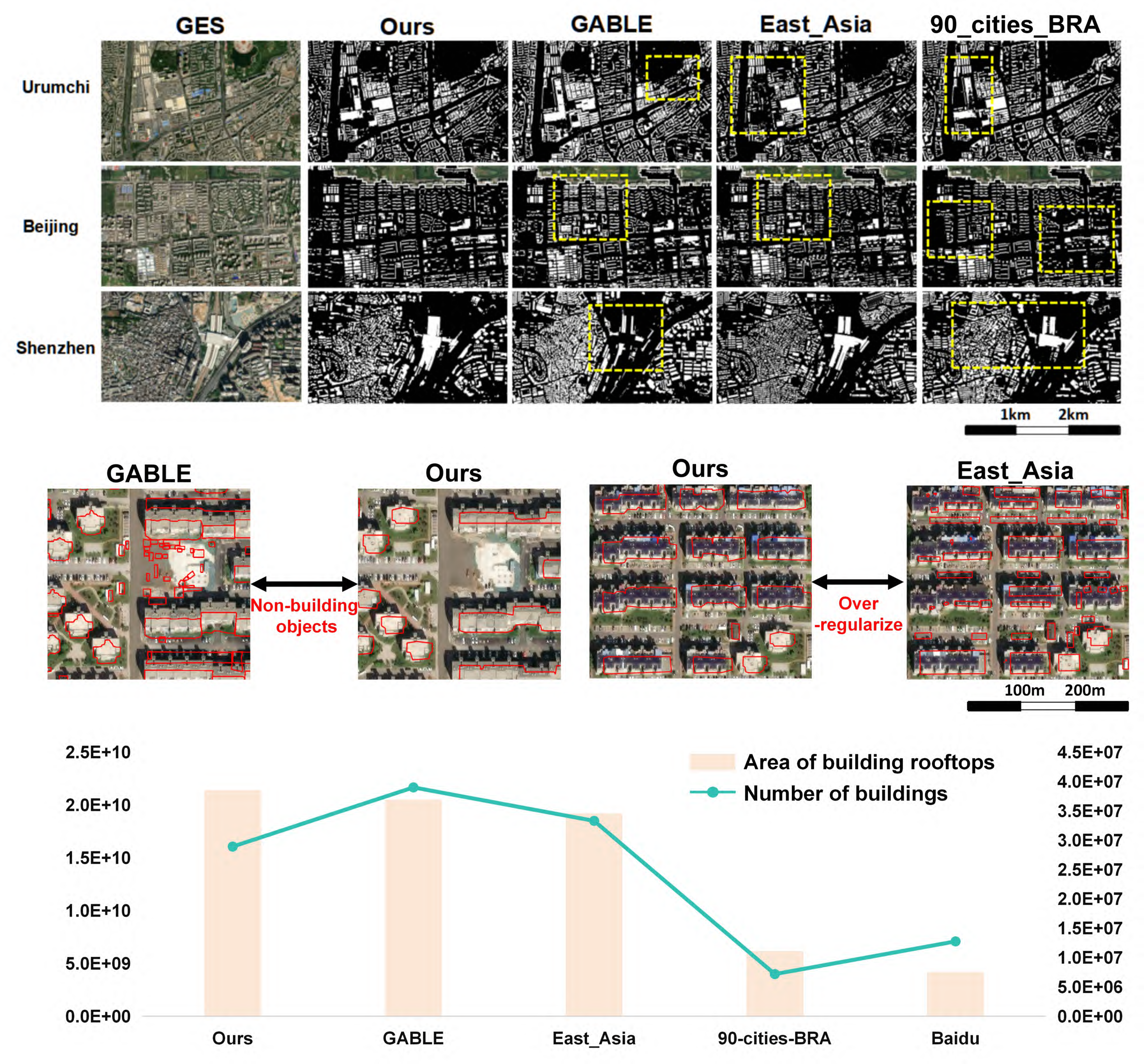}  
\caption{\textbf{Visualization and Comparison of building rooftops with different building rooftops data in the boundary of natural cities we used. } \small{The limitation of CBRA using low-resolution data for super-resolution segmentation is that the building can't achieve vector output, so this paper omits the comparison with CBRA.}}
\label{fig:stream}
\end{figure}
\FloatBarrier 
From the recognition results (Figure 9), our data product demonstrates superior accuracy and completeness compared to existing building footprint datasets. Given the increased focus on spatial cities, we conducted a comparative analysis of building areas identified by different data products. Our findings indicate that our approach identifies a greater number of buildings in spatial cities. This outcome is expected, as our methodology involved the manual interpretation and comparison of remote sensing images across all spatial cities. The Douglas-Peucker algorithm with an empirical threshold used for vectorizing the contours distinguishes well between buildings and non-building objects such as cars, compared to GABLE and East\_Asia buildings datasets. Additionally, our method aligns better with visual interpretation than the further black-box post-processed East\_Asia dataset (using GAN), which tends to over-regularize, although some untreated roofs might appear less aesthetically pleasing due to retaining more segmented shapes. Regarding the recognition of buildings in remote areas, our climate zone annotations significantly enhanced the accuracy for some special buildings, such as large religious structures, compared to GABLE, East Asia, and 90\_cities\_BRA datasets. The total recognized building area in existing studies is similar, but the building count varies greatly for two main reasons: first, whether multiple roofs of a single building are identified separately, and second, whether small structures are mistakenly identified as buildings.

\subsubsection{Building height}
Firstly, the comparison of the height partition models (A, B, C, D, E, representing models trained according to administrative levels) and the combination model, integrated through 100 training iterations using the Bootstrap Aggregated XGBoost method, is presented (see Supplementary Table 4 for details). It can be observed that, except for the partition model trained on building data of level E, the accuracy metrics of the other partition models surpass those of the combination model. Notably, the R² values for the first two partition models exceed 0.8, with building height prediction errors less than 6 meters. This discrepancy can be explained by the substantial variation in construction investment intensity across different administrative levels. Buildings in higher-level cities are more influenced by socioeconomic factors, making their heights relatively harder to predict, as reflected in the higher R² values for building predictions in lower-level cities. Using SVIs, we audited 2500 building heights through manual observation of floor counts and further estimated their heights. Regression analysis against the dataset resulted in an R² of 0.72, indicating accurate identification of building heights in most cases.

Secondly, we compared different products through visualization and height segmentation, focusing on Baidu and GABLE datasets (Figure 10). Baidu data is used as the ground truth for Chinese building heights in most studies, including ours, which indicates the accuracy of our data product through the similarity in distributions. Furthermore, GABLE, the only product that identifies the height of all buildings from optical images across China, provides RMSE values for height intervals of 0-10, 10-30, 30-50, 50-100, and 100-500 meters. By categorizing our data according to these intervals, we found that our product exhibits lower RMSE values in the intervals below 50 meters. Based on statistics of Baidu data we used, 98\% of building heights are below 50 meters. Additionally, according to the 2020 Chinese Census, residential buildings with more than 10 stories (roughly equivalent to building height higher than 30m) account for only 1\%. Therefore, these results indicate that our product achieves better height prediction results for the vast majority of buildings.

The Relative Error range/Absolute Error range for the combination model, partition model C, partition model D, and partition model E is lower than partition models A and B (see Figure S5). This indicates that the overall accuracy decreases with higher city administrative levels, but the uncertainty in estimates due to different data partitioning methods and model parameters is lower, resulting in more stable estimates. The relative error curves of the four stable estimation models start to rise sharply around 20 meters, indicating that the deviation in building height estimates increases with the true height value, while the relative error for buildings below 20 meters is relatively low. The uncertainty is reduced in the partition models compared to the combination model (especially in the comparison between models C/D/E and the combination model).

\begin{figure}[!htbp] 
\centering
\includegraphics[width=1\textwidth]{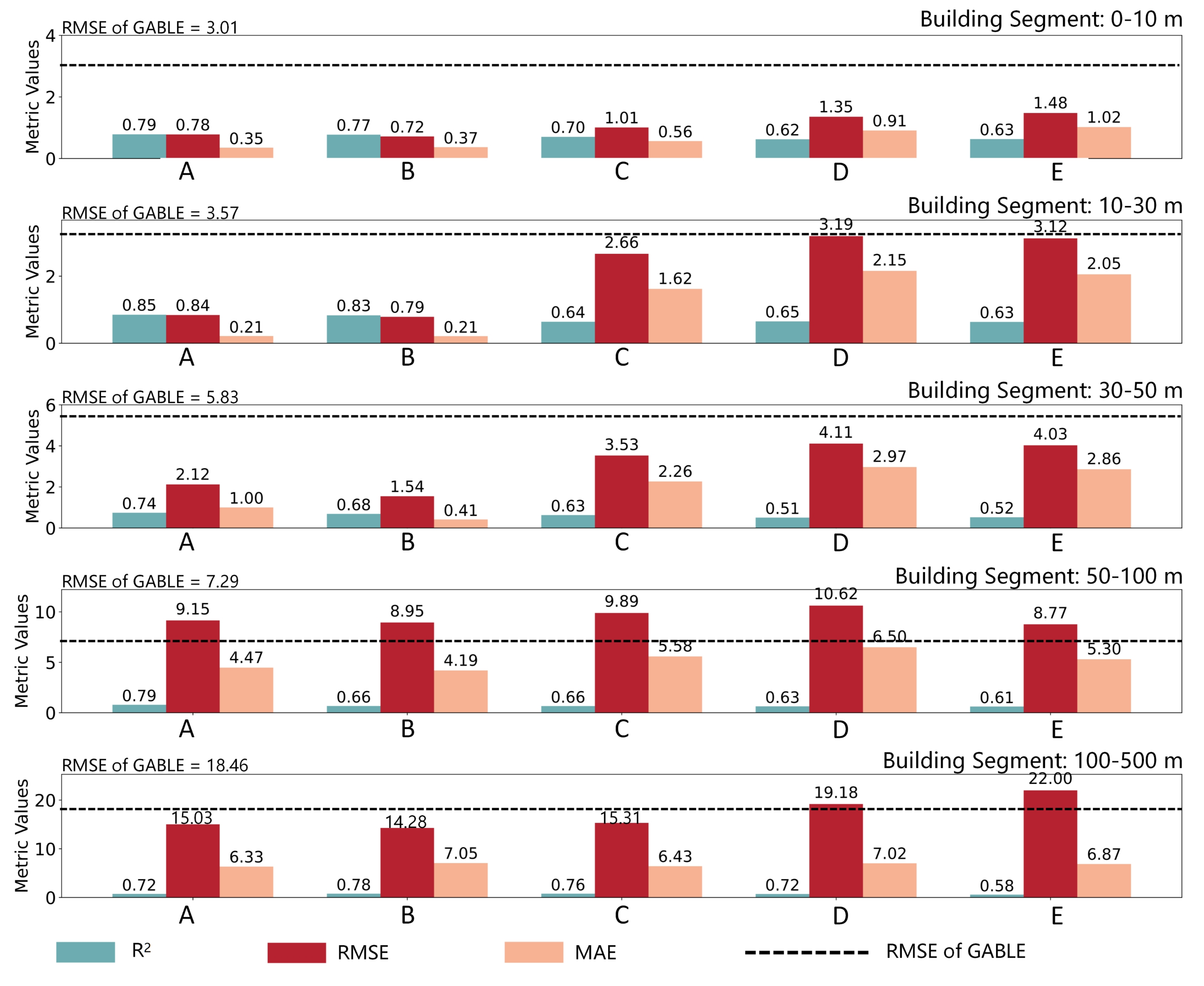}  
\caption{\textbf{Visualization and Comparison building height model with GABLE.}A, B, C, D, E stand for the height model that is trained according to administrative levels (see Table 4).}
\label{fig:stream}
\end{figure}
\FloatBarrier 

\newpage
\subsection{Model evaluation and comparison for Indicative attributes}
\subsubsection{Building function}
The results clearly indicate that the partition models outperform the combined model, with Model A showing superior accuracy across multiple functional categories in Figure S6. Notably, the precision for residential function identification is higher, with an F1-score approaching 0.90. Other functions, such as office buildings, have slightly lower identification precision, with F1-scores nearing 0.80. In contrast, the performance of the commercial and public service models is suboptimal, with F1-scores around 0.5. This discrepancy may be attributed to the varying sample sizes of different building types. Using SVI data, we verified the functions of 2500 buildings through manual auditing, observing details such as building names and architectural styles to determine their functions. Comparing these observations with the dataset, we found that 88\% of buildings were accurately classified in terms of their functions. This indicates that the functional purposes of the majority of buildings were correctly identified through our methodology.

Furthermore, we compared our study with recent research that identified the primary functions of buildings by calculating the geometric features of building coverage, distances to adjacent objects, and the kernel density of POIsnts of interest (POIs) \citep{LZ2023}. Firstly, unlike the aforementioned study, which focused on three urban agglomerations in China (Beijing-Tianjin-Hebei, Yangtze River Delta, and Pearl River Delta), our research encompasses buildings on a national scale. Secondly, in terms of model accuracy, our functional prediction model's accuracy is slightly lower than that of the recent study (average accuracy of 0.93), likely due to the latter's more limited scope. Nonetheless, the performance of our models across different functions aligns with the findings of that study, exhibiting strong performance in residential functions (average accuracy of 0.97) while demonstrating weaker performance in commercial (average accuracy of 0.63) and public service functions (average accuracy of 0.67).

\subsubsection{Building quality and age}
Judging from the model evaluation, the accuracy of building quality depends on the accuracy of the Yolo-v8 model. According to \citet{CJ2023} and \citet{LY2024}, the identification accuracy of various building quality categories is as follows: "Buildings with damaged facades" (83.4\%), "illegal/temporary buildings" (71.6\%), “Graffiti/illegal advertisement” (80.7\%), “Stores with poor facades” (89.8\%), “Buildings with unkempt facades” (79.9\%), and “Stores with poor signboards” (84.6\%). The accuracy of building age depends on the accuracy of GAIA data. According to \citet{GP2020}, the mean overall accuracy over years of GAIA is higher than 90\%. 

We validate this method using national housing trade data (3,771,892 house rent and 608,984 community records with coordinates and construction years with the coverage of 2,490 spatial cities) from Anjuke.com, one of China's largest real estate trading platforms. We found that the years in the housing trade data of all provinces were significantly positively correlated with the presentation we recognized (\textit{P} < 0.05), but because the housing price data appeared in the form of one POIsnt in a community, we could not completely validate our data.

Therefore, we finally used SVIs to manually mark and validate the quality and age of 2500 buildings using SVIs through manual auditing, assigning quality problems severity scores from 0 to 6, where higher scores indicate more severe issues of building quality. The correlation analysis with existing results yielded an $R^2$ value of 0.78, indicating accurate identification of quality issues in the majority of buildings. Regarding the building age, we divided the building age in manual labeling into five categories (time division POIsnts are 1985, 1990, 2000, 2010, and 2018), and checked whether the identified building age conforms to the observed true value of the category, and found that 82\% of the buildings are consistent. This shows that most building ages are accurately classified. The reason we can't make statistics directly according to the year of building completion is that, firstly, we can't determine the year of completion just by visual inspection; and secondly, the experiment assumes that the expansion of impervious surfaces is synchronous with building construction age, whereas in reality, some buildings are demolished, and others appear before the impervious surfaces are established.

\section{Usage Notes}
The distribution of buildings in spatial cities within administrative boundaries, assessed by the total rooftop area, average building height, and building stock volume, indicates that the spatial distribution of building stock in China is predominantly concentrated in the Beijing-Tianjin-Hebei region, the Central Plains, the Yangtze River Delta, and the Pearl River Delta, with obvious differences between the two sides of the Hu Weiyong line. The building rooftop area demonstrates a pattern of higher values in the east and lower values in the west while building heights exhibit a spatial distribution of higher values in the south and lower values in the north. Notably, central and western provinces, such as Tibet and Qinghai, have a small building stock and low average heights, showing a significant disparity compared to eastern coastal regions like Shandong, Jiangsu, and Guangdong Province (Figure 11).

Based on the inferred results of building functions, the current building stock in China is predominantly composed of residential buildings (74.3\%), followed by public service facilities (10.6\%) and office buildings (9.7\%), with industrial buildings having the lowest proportion (0.39\%). Figure 12 visualizes the quantity and proportion of different types of buildings across various provinces. All building types meet the law of high in the east and low in the west in terms of quantity. The eastern regions exhibit a higher proportion of office and industrial buildings. Contrary to common expectations, the proportion of public service buildings in the western regions is significantly higher than in the eastern regions (Figure 12).

\begin{figure}[!htbp] 
\centering
\includegraphics[width=1\textwidth]{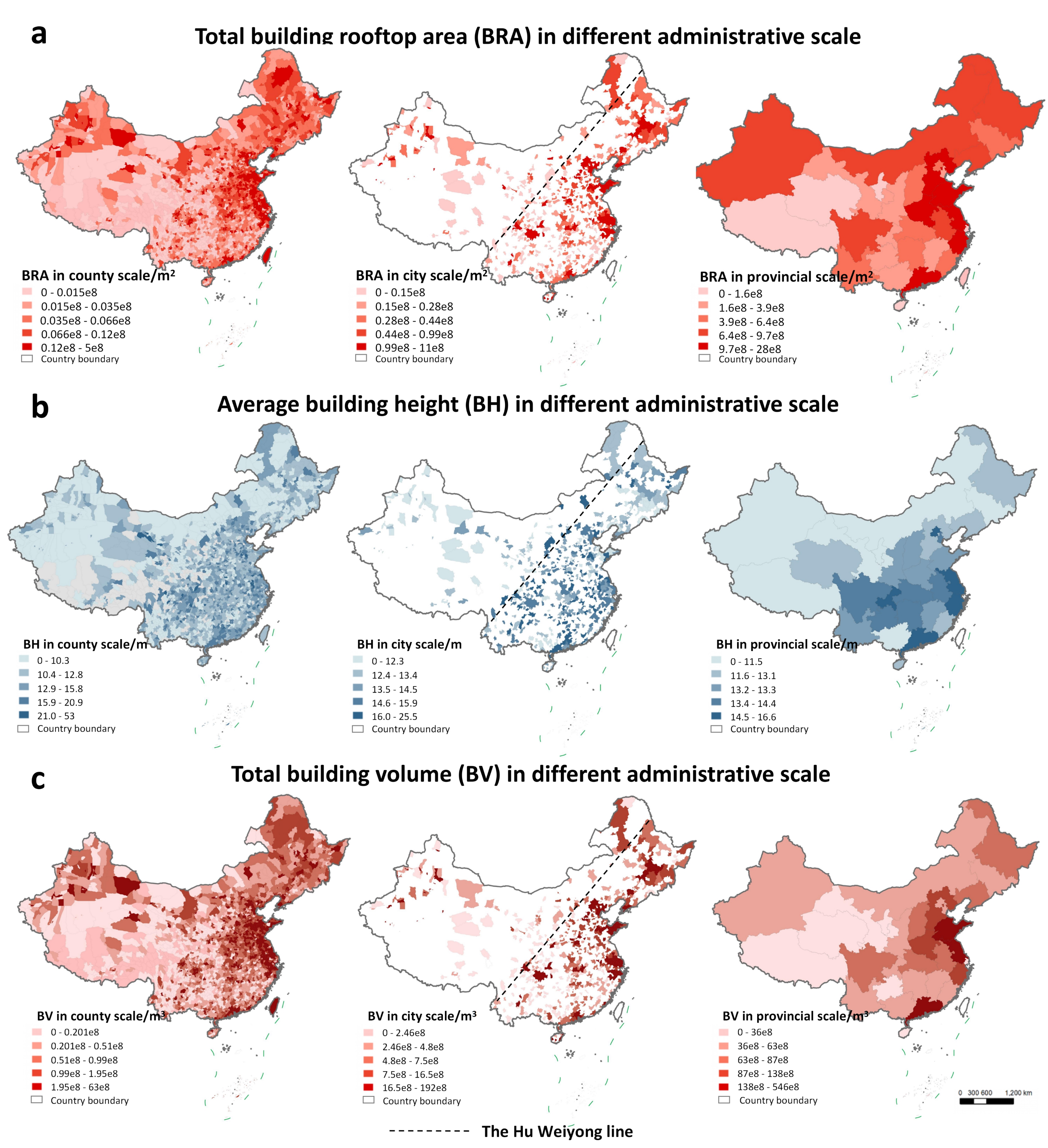}  
\caption{\textbf{Visualization of 3D building data.}}
\label{fig:stream}
\end{figure}
\FloatBarrier 

\begin{figure}[!htbp] 
\centering
\includegraphics[width=1\textwidth]{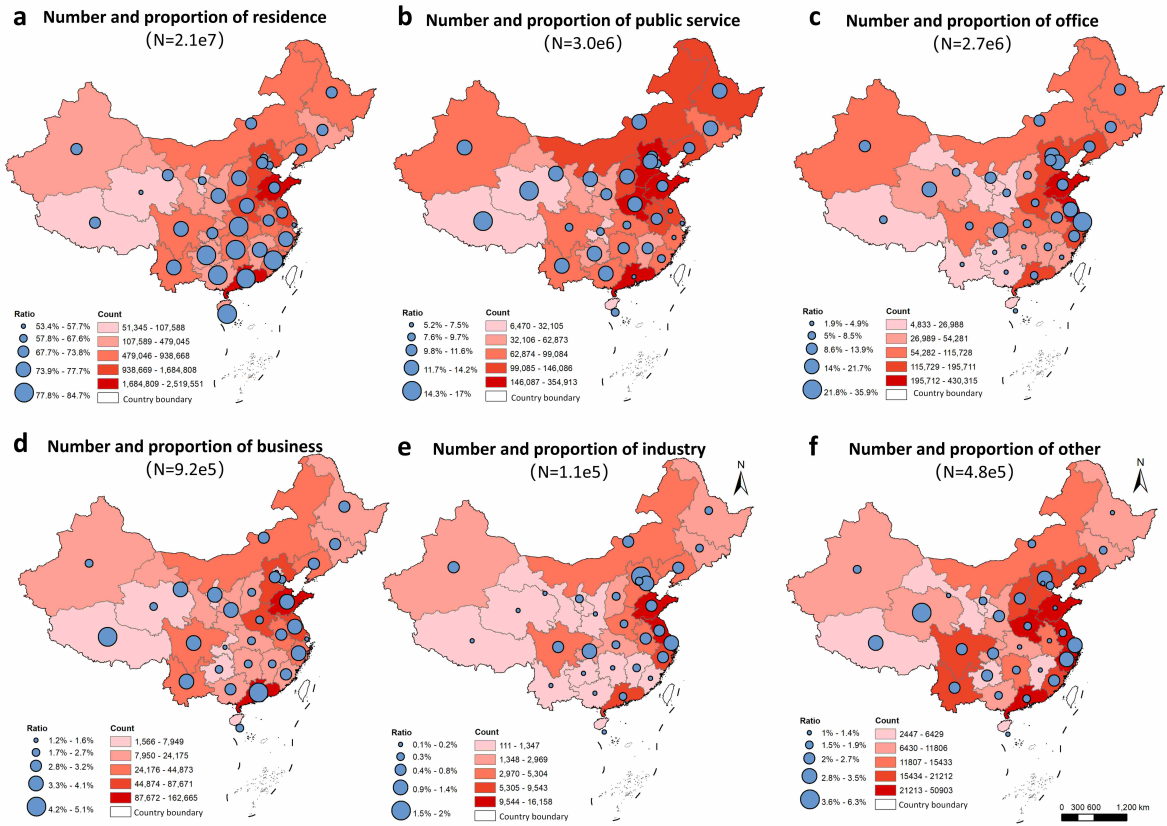}  
\caption{\textbf{Visualization of building function.}}
\label{fig:stream}
\end{figure}

Figure 13 shows the spatial distribution pattern and quantitative statistical trend in the quality of buildings along the street (accounting for 29.4\% of all the buildings) evaluated by street view imagery, which revealed the significant regional differences in building quality in China. The most severe issues in the buildings along the streets are “Illegal/temporary buildings” (47.0\%) and “unkempt store facades” (27.1\%), followed by “Graffiti/illegal advertisement” (12.9\%), “Stores with poor signboards” (9.4\%), “Buildings with unkempt facades” (6.5\%) and “Buildings with damaged facades” (2.4\%). The building quality is notably higher in the eastern and southern regions, while the northeastern, northwestern, and southwestern regions have lower building quality. The northeastern and northwestern regions face severe “Buildings with damaged facades” problems. In contrast, the primary issue in the eastern provinces is “Graffiti/illegal advertisement”.

\newpage
\begin{figure}[!htbp] 
\centering
\includegraphics[width=1\textwidth]{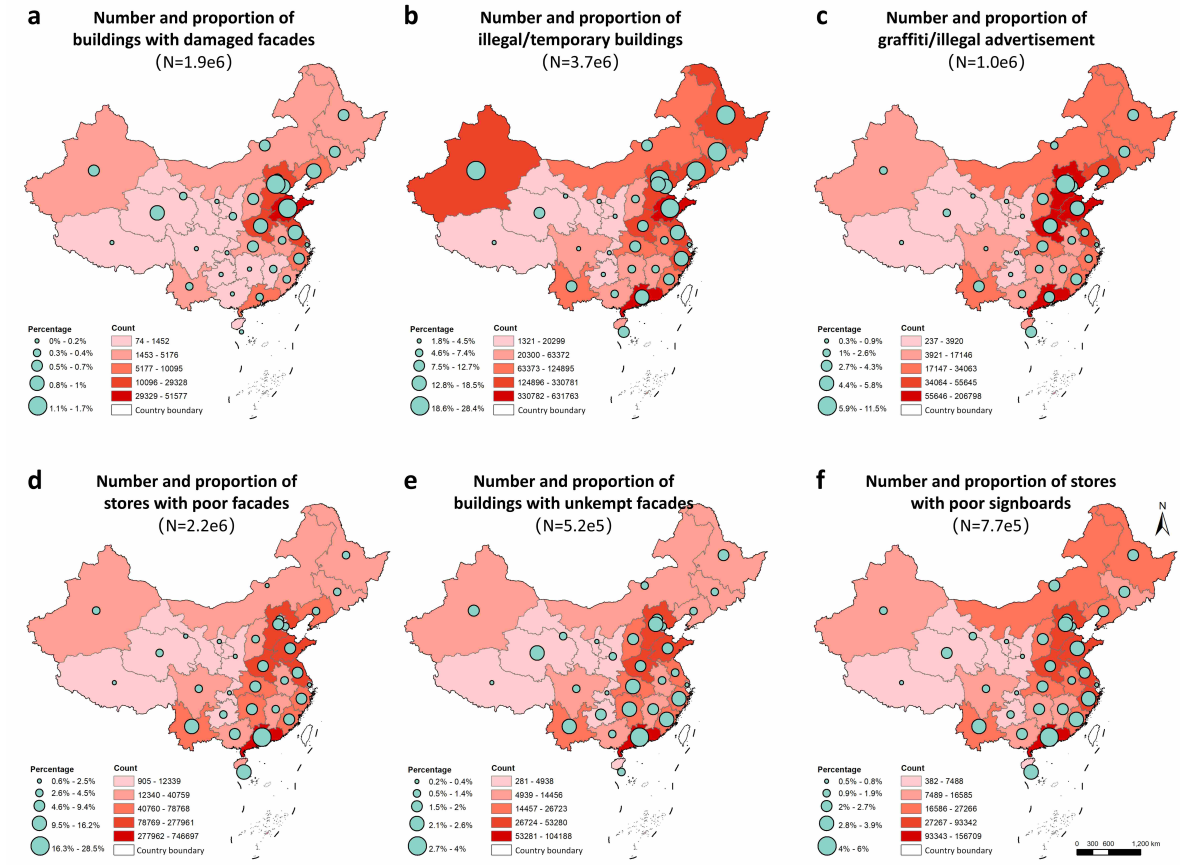}  
\caption{\textbf{Visualization of building quality.} \small{The six building quality scores—(“Buildings with damaged facades”(DamFa), “Illegal/temporary buildings”(IllBu), “Graffiti/illegal advertisement”(IllAd), “Stores with poor facades”(UnkSt), “Buildings with unkempt facades”(UnkFa), and “Stores with poor signboards”(PooSi)—have three modes: No observation POIsnts in the buffer zone; No street view images in the observation POIsnts for that year; a numeric value (0-1) represents the quality score, with higher values indicating more severe issues. Building quality is the total score (0-6) for the six quality issues in the most recent year, which has the numeric value for six building quality scores, indicating the severity of building quality problems. Building quality issues are analyzed by administrative divisions.}}
\label{fig:stream}
\end{figure}

By visualizing the number of buildings constructed in different provinces and the proportion of these buildings relative to the total existing buildings in each province over the past 40 years, it can be observed that the number of buildings in urban China has increased by about 5 million buildings per decade approximately from 1985 to 2015, with an increase of about 5 million buildings per decade. From 2005 to 2015, the number of urban buildings increased the most, with an increase of nearly 10 million. However, the past five years (2019 to present) have been the lowest period of building expansion in urban area, with an increase of 3.4 million buildings. Shandong and Guangdong Province, among others, have experienced the fastest growth (Figure 14).

\begin{figure}[h!] 
\centering
\includegraphics[width=1\textwidth]{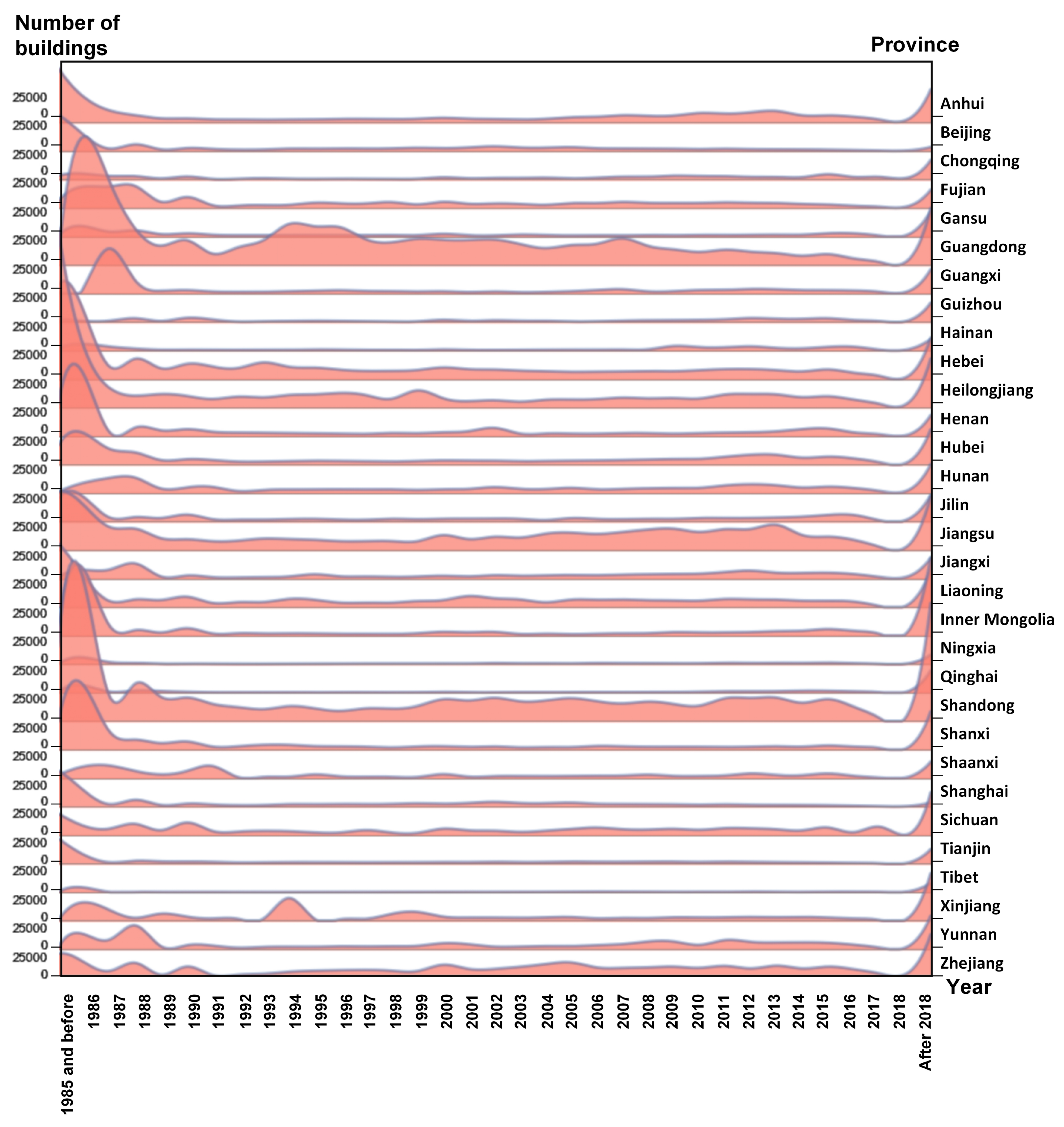}  
\caption{\textbf{Visualization of building age.} \small{“AF2018” represents building data that were built after 2018.}}
\label{fig:stream}
\end{figure}

\newpage
\section{Discussion}
This study has extracted the world's first national-scale multi-attribute building dataset at the level of building instances, encompassing attributes such as roof type, height, function, age, and quality. Through a combination of manually labeled remote sensing images, deep learning, manually created multi-scale building features, and Bootstrap Aggregated XGBoost, we have improved the accuracy of rooftop and height extraction of urban 3D building data compared to similar products. This dataset is one of only two that covers the entire natural urban area of China. By integrating tens of millions of street view and geographic foundational data, GAIA remote sensing data products, we generated building functions, ages, and qualities based on height data, uncovering the spatiotemporal distribution patterns of these building attributes on a national scale. Additionally, we verified our attribute predictions through manually labeled street views and multi-source data. The methods, datasets, and findings of this study lay an important foundation for future urban renewal and redevelopment, and will effectively promote the achievement of SDG goals.

This study demonstrates that the combination of various remote sensing datasets, street view datasets, and other spatial data can accurately estimate the full-chain attribute data of buildings, including rooftop, height, function, age, and quality. By experimenting with multi-source heterogeneous data and different methods, we revealed the primary data sources and validation sources for obtaining each building attribute. Consequently, our model can depict the urban physical and social landscape of 3D buildings, analyze the structural characteristics of different areas (such as cities of different levels), and examine the differences and fairness among various geographical features. Our proposed 3D data of roofs and heights exhibit high accuracy (the average overall accuracy of rooftops exceeds 0.9, with a significant improvement in height accuracy for buildings below 50 meters). In terms of $R^2$, RMSE, and MAE, it outperforms other instance-level 3D building models. Furthermore, for the integrated learning models of building height and function, this study reveals the most important predictive feature variables, which are crucial for understanding urban development and evolution.

This study further uncovers some unique patterns of building distribution in China through the creation of this dataset. Firstly, in the 3D building data, our roof and height data exhibit spatial distribution patterns similar to existing data products, showing that the building stock is generally higher in the south than in the north and higher in the east than in the west of China. Compared to previous studies, this study further reveals the building stock distribution at different scales (including county, city, and provincial levels) and its association with the Hu Line. This may provide significant updates to previous research conclusions that used census data for large-scale carbon emission measurements. Secondly, for building function, quality, and age, we have provided the first nationwide quantitative analysis of different attributes at the building scale. We found that the building stock in China predominantly has residential functions, revealing differences in building function distribution across regions and reflecting the spatial distribution differences of urban social structures. Additionally, using nationwide street view images over several years, we uncovered the spatial distribution patterns of building quality for nearly one-third of street-facing buildings, identifying illegal/temporary buildings as the most severe building quality issue in China. The regional differences in building quality proportions between central and western China will guide the next steps in urban renewal plans. Finally, we presented the spatial distribution of the ages of all buildings nationwide, showing the annual growth rate of the building stock in China over the past forty years.

The generated dataset of 3D building attributes, including roof type and height, offers significant opportunities for advancing urban studies. This data can facilitate the classification of different settlement types (e.g., residential, commercial, industrial zones), aiding in the analysis of social and environmental impacts of urban areas \citep{TH2018}. Additionally, information on building height and footprint is essential for disaster risk assessment, enabling better evaluation of building vulnerabilities to natural hazards such as floods, storms, and earthquakes \citep{DS2018, KE2020, PD2020}. The height data is also crucial for urban heat island modeling, providing insights into the impact of urban morphology on local climates and heat stress \citep{LA2015}. Overall, the 3D attributes enhance our understanding of urban structure and its influence on various environmental conditions \citep{SK2009}.

The additional attributes of building function, age, and quality further expand the dataset's utility. Detailed function data reveals the spatial distribution of urban social structures and highlights areas dominated by residential, commercial, or industrial functions. This information is crucial for urban planning and socioeconomic analysis. The age attribute allows for the examination of historical urban development and growth patterns, facilitating studies on urban expansion and policy impacts over time \citep{ZY2017}. Building quality data is vital for assessing regional disparities and guiding urban renewal efforts, particularly in identifying areas with significant proportions of illegal or temporary buildings \citep{RM2018}. Together, these attributes enable a comprehensive analysis of urban dynamics and support cross-disciplinary research in public health, economics, and environmental science.

Based on our research framework, future studies should focus on the supplementation and maintenance of comprehensive building attributes, such as structural integrity, occupancy status, and energy efficiency. Continuous updating and accuracy of these attributes are essential to maximizing the dataset's utility for urban planning, policy-making, and sustainable development initiatives. Moreover, the integration of new data sources, including IoT sensors, real-time satellite imagery, and crowdsourced data, will provide richer, more dynamic datasets. Leveraging advanced machine learning techniques, such as deep learning and ensemble methods, will be crucial for refining predictions and expanding the scope of this research. These enhancements will enable more precise modeling of urban environments, support disaster resilience planning, optimize resource allocation, and contribute to more sustainable and livable cities. Future research should also explore the socioeconomic impacts of urban morphology, examining how variations in building attributes affect factors such as property values, demographic trends, and social equity. This holistic approach will provide deeper insights into urban dynamics and foster data-driven decision-making processes across multiple domains.

\section*{Code availability}
The CMAB dataset was created using Python 3.9.7 as well as ArcGIS 11.2, and the code and datasets can be available at Mendeley after the publication of the paper. It takes 10 minutes to extract 1000 square kilometers of buildings with NVIDIA 3070 using our methodology.

\section*{Author contributions statement}
YZ: Writing - original draft, Methodology, Data curation, Software, Visualization, Validation, Formal Analysis; HZ: Writing - original draft, Visualization, Investigation, Validation; YL: Conceptualization, Funding acquisition, Project Administration, Supervision, Resources, Writing – Review \& Editing.

\section*{Acknowledgements} 
Thanks to Yue Ma, Enjia Zhang, Xinyu Wang and Junhao Xia for their contributions to the original data and suggestions.

\section*{Competing interests}
All authors declare no financial or non-financial competing interests.

\newpage 

\clearpage 
\section*{Supplemental Material}
Overview: This Supplementary Information provides additional information about the methods and analysis about this work.\\
\textbf{Supplementary Notes}\\
Supplementary note 1. OCRNet for building rooftop segmentation\\
Supplementary note 2. Feature calculation of buildings\\
\textbf{Supplementary Tables}\\
Supplementary Table 1. Baidu AOI classification.\\
Supplementary Table 2. Construction of building functional attributes.\\
Supplementary Table 3. Description of the attributes’ fields in data products.\\
Supplementary Table 4. Comparison of height partition model and combination model.\\
\textbf{Supplementary Figures}\\
Supplementary Figure 1. Time and area distribution of GES imagery for spatial cities.\\
Supplementary Figure 2. Resampling process of the high-density urban area with label enhancement.\\
Supplementary Figure 3. Statistics and preprocessing of Baidu building dataset.\\
Supplementary Figure 4. The samples of building feature calculation.\\
Supplementary Figure 5. Uncertainty expressed in the Relative Error and Absolute Error values of the aggregated method for partition model and combination model.\\
Supplementary Figure 6. Accuracy of building function model.\\

\renewcommand{\theequation}{\arabic{equation}} 
\setcounter{equation}{0} 
\subsection*{Supplementary Notes}
\subsubsection*{Supplementary note 1. OCRNet for building rooftop segmentation}
Specifically, considering computational load and efficiency, this study utilized remote sensing images from 3667 natural cities, with each city divided into multiple 500×500 pixel grid slices $I_i$. The backbone network employed HRNet\_w48 for feature extraction. Assuming the feature extractor is $\varphi(\cdot)$, for the input slice $I_i$, the backbone network extracts the feature map $F_i$ \ref{eq:Fi} and calculates the Object Representation $R_i$ \ref{eq:Ri} for each pixel in the slice:
\begin{align}
F_i &= \varphi(I_i) \label{eq:Fi} \\
R_i &= \theta(F_i) \label{eq:Ri}
\end{align}

In this context, the object representation module $\theta(\cdot)$ computes the relevance of each pixel concerning the two categories: building roofs and background, through an attention mechanism. This calculation involves the Contextual Representation $C_i$ \ref{eq:Ci}, which integrates global contextual information:
\begin{align}
C_i &= \mu(F_i, R_i) \label{eq:Ci}
\end{align}

This includes the context representation module $\mu(\cdot)$, which computes contextual information by integrating object representations with feature maps. Finally, a classification head $S_i$ \ref{eq:Si} is used to transform the contextual representations into segmentation results for building roofs:
\begin{align}
S_i = \rho(C_i) \label{eq:Si}
\end{align}

This includes the classification head $\rho(\cdot)$, which is used to convert the contextual representations into classification results for each pixel. During the training process, the model is optimized using a cross-entropy loss function. Assuming the labels are $Y_i$, where N is the number of training samples of annotated building roof segments, the loss function $L_i$ is defined as:
\begin{equation}
L_i = \frac{1}{N} \sum_{i=1}^{N} \text{CrossEntropy}(S_i, Y_i)
\label{eq:Li}
\end{equation}

\subsubsection*{Supplementary note 2. Feature calculation of buildings}
We take the calculation methods of "is building along the street or not" and "distance to urban functional center" as examples to illustrate the rationality of our feature calculation (Figure S4). To determine whether a building is adjacent to a street, it is necessary to reconstruct a reasonable boundary from the collection of buildings within the same block, along with an appropriate street setback constraint. We define the concave hull as the area occupied by the set of building vertices. The alpha-shape algorithm in Python approximates this area by creating a concave polygon. Buildings intersecting with the concave hull boundary (alpha=0.01) and within a certain distance threshold (street setback) from the street ensure that non-street-facing recessed buildings are excluded. Through multiple trials, a street setback constraint of 100m and an alpha value of 0.01 was found to be suitable empirical thresholds for determining if a building is adjacent to a street.

Regarding to calculation of the distance to urban functional centers, this study references related research (Li, 2018) to obtain a two-dimensional kernel density analysis surface of different natural cities using the boundaries of physical cities as a mask based on 13,884,254 POIs (POIsnt of Interest) POIsnts nationwide. In the surface, the height value of each POIsnt represents the density estimate at that location. The kernel density estimate ${f}(x, y)$ \ref{eq:kernel_density} \ref{eq:kernel_function} at a POIsnt $(x,y)$ using a set of $n$ POIs POIsnts $\{(x_i, y_i)\}_{i=1}^{n}$ is given by:
\begin{align}
\hat{f}(x, y) &= \frac{1}{nh^2} \sum_{i=1}^{n} K \left( \frac{x - x_i}{h}, \frac{y - y_i}{h} \right), \label{eq:kernel_density} \\
K(u, v) &= \frac{1}{2\pi} e^{-\frac{u^2 + v^2}{2}} \label{eq:kernel_function}
\end{align}

where $K$ is the Gaussian kernel function, and $h$ is the bandwidth parameter. The local maxima POIsnts of the surface are detected by calculating the Hessian matrix. The Hessian matrix $H$ \ref{eq:hessian} at a POIsnt $(x,y)$ on the density surface ${f}(x, y)$ is given by: 
\begin{equation}
H_{ij} =
\begin{vmatrix}
\frac{\partial^2 \hat{f}}{\partial x^2} & \frac{\partial^2 \hat{f}}{\partial x \partial y} \\
\frac{\partial^2 \hat{f}}{\partial y \partial x} & \frac{\partial^2 \hat{f}}{\partial y^2}
\end{vmatrix}
\label{eq:hessian}
\end{equation}

A POIsnt $(x,y)$ is a local maximum if the Hessian matrix is negative definite. The cluster of POIsnts around $(x,y)$ forms the multifunctional center of the city. Then we calculate the distance from the centroid of each building to the nearest multifunctional center as a representation of this feature. To determine if a POIsnt $(x,y)$ is a local maximum, we check the definiteness of the Hessian matrix $H$. This requires all eigenvalues of to be negative. The Hessian matrix is negative definite if \ref{eq:example}:
\begin{equation}
\lambda_1 < 0 \text{ and } \lambda_2 < 0
\label{eq:example}
\end{equation}

\clearpage 
\subsection*{Supplementary Tables}
\subsubsection*{Supplementary Table 1. Baidu AOI classification.}
\begin{longtable}{|c|p{4cm}|p{9cm}|}
\hline
\textbf{Num} & \textbf{Primary type} & \textbf{Secondary type} \\
\hline
\endfirsthead
\hline
\textbf{Num} & \textbf{Primary type} & \textbf{Secondary type} \\
\hline
\endlastfoot
1  & Restaurants           & Chinese Restaurants, Foreign Restaurants, Snack and Fast Food Shops, Cake And Dessert Shops, Cafés, Tea Houses, Bars, Others \\
\hline
2  & Hotels                & Star-Rated Hotels, Budget Hotels, Apartment Hotels, Homestays, Others \\
\hline
3  & Shopping              & Shopping Centers, Department Stores, Supermarkets, Convenience Stores, Home Improvement and Building Materials, Electronics and Appliances, Retail Shops, Markets, Others \\
\hline
4  & Lifestyle Services    & Communication Service Halls, Post Offices, Logistics Companies, Ticket Offices, Laundry Services, Printing and Copying Shops, Photo Studios, Real Estate Agencies, Public Utilities, Repair Services, Domestic Services, Funeral Services, Lottery Sales POIsnts, Pet Services, Newsstands, Public Toilets, Dedicated Bike Paths Rest Stations, Others \\
\hline
5  & Beauty and Wellness   & Beauty Salons, Hair Salons, Nail Salons, Body Treatments, Others \\
\hline
6  & Tourist Attractions   & Parks, Zoos, Botanical Gardens, Amusement Parks, Museums, Aquariums, Beaches, Historical Sites, Churches, Scenic Areas, Tourist Spots, Temples, Others \\
\hline
7  & Leisure and Entertainment & Resorts, Farmhouses, Cinemas, KTVs, Theaters, Dance Halls, Internet Cafés, Gaming Venues, Spa and Massage Centers, Leisure Squares, Others \\
\hline
8  & Sports and Fitness    & Sports Venues, Extreme Sports Facilities, Fitness Centers, Others \\
\hline
9  & Education and Training & Universities, Secondary Schools, Primary Schools, Kindergartens, Adult Education, Parent-Child Education, Special Education Schools, Study Abroad Agencies, Research Institutes, Training Institutes, Libraries, Science and Technology Museums, Others \\
\hline
10 & Culture and Media     & News And Publishing, Broadcasting and Television, Art Troupes, Art Galleries, Exhibition Halls, Cultural Palaces, Others \\
\hline
11 & Healthcare            & General Hospitals, Specialized Hospitals, Clinics, Pharmacies, Health Examination Centers, Sanatoriums, Emergency Centers, Disease Control Centers, Medical Equipment, Healthcare Services, COVID-19 Testing Sites, COVID-19 Vaccination Sites, Risk POIsnts, Temporary Hospitals, Fever Clinics, Others \\
\hline
12 & Automotive Services   & Car Sales, Car Repairs, Car Detailing, Auto Parts, Car Rentals, Vehicle Inspection Centers, Others \\
\hline
13 & Transportation Infrastructure & Airports, Railway Stations, Subway Stations, Subway Lines, Long-Distance Bus Stations, Bus Stops, Ports, Parking Lots, Parking Areas, Parking Spaces, Gas and CNG Stations, Service Areas, Toll Stations, Bridges, Charging Stations, Roadside Parking, Standard Parking, Pickup And Drop-Off POIsnts, Electric Bicycle Charging Stations, Highway Rest Areas, Others \\
\hline
14 & Finance               & Banks, ATMs, Credit Unions, Investment and Financial Services, Pawn Shops, Others \\
\hline
15 & Real Estate           & Office Buildings, Residential Areas, Dormitories, Internal Building Units, Others \\
\hline
16 & Corporations and Enterprises & Companies, Industrial Parks, Agricultural and Horticultural Farms, Factories and Mines, Others \\
\hline
17 & Government Institutions & Central Institutions, Government Agencies at All Levels, Administrative Units, Public Security and Judicial Institutions, Foreign Affairs Agencies, Political Parties and Groups, Welfare Institutions, Political Education Institutions, Social Organizations, Democratic Parties, Resident Committees, Others \\
\hline
18 & Entrances and Exits   & Highway Exits, Highway Entrances, Airport Exits, Airport Entrances, Station Exits, Station Entrances, Gates (Note: Doors of Buildings Or Building Complexes), Parking Lot Entrances And Exits, Bicycle Highway Exits, Bicycle Highway Entrances, Bicycle Highway Entrances And Exits, Parking Lot Exits, Parking Lot Entrances, Others \\
\hline
19 & Natural Features      & Islands, Mountains, Water Bodies, Others \\
\hline
20 & Administrative Landmarks & Provinces, Provincial Cities, Prefecture-Level Cities, Districts and Counties, Commercial Areas, Townships, Villages, Others \\
\hline
21 & Addresses             & Address POIsnts, Others \\
\hline
22 & Roads                 & Highways, National Roads, Provincial Roads, County Roads, Township Roads, Urban Expressways, Urban Main Roads, Urban Secondary Roads, Urban Side Roads, Ferry Lines, Intersections, Others \\
\hline
23 & Railways              & Railways, Subways/Light Rail, Maglev Trains, Trams, Intercity Rapid Rail, Others \\
\hline
24 & Administrative Boundaries & International Boundaries, Established National Boundaries, Undetermined National Boundaries, Hong Kong And Macau Boundaries, South China Sea Boundary Lines, Established Provincial Boundaries, Undetermined Provincial Boundaries, Coastlines, Others \\
\hline
25 & Other Linear Elements & Bridges, Tunnels, Administrative Imaginary Lines, Water Area Imaginary Lines, Green Space Imaginary Lines, Island Imaginary Lines, Epidemic Control Areas, Others \\
\hline
26 & Administrative Divisions & World-Class, National-Level, Provincial-Level, Municipal-Level, District and County-Level, Hotspot Areas, Built-Up Areas, Smart Areas, Others \\
\hline
27 & Water Bodies          & Dual-Line Rivers, Lakes and Marshes, Oceans, Others \\
\hline
28 & Green Spaces          & Green Parks, Golf Courses, Islands, Green Belts, Airports, Airport Roads, Others \\
\hline
29 & Labels                & Continent Labels, Ocean Labels, Sea Area Labels, Water Body Labels, Island Labels, Non-Water Body Labels, Others \\
\hline
30 & Bus Routes            & Regular Daytime Bus Routes, Subways/Light Rail, Trams, Airport Buses (To Airport), Airport Buses (From Airport), Airport Buses (Between Airports), Tourist Bus Routes, Night Buses, Ferries, Express Buses, Slow Buses, Airport Express (To Airport), Airport Express (From Airport), Airport Rail Transit Loops, Others \\
\hline
\end{longtable}
\FloatBarrier 

\subsubsection*{Supplementary Table 2. Construction of building functional attributes.}
\begin{table}[ht]
\centering
\small
\resizebox{\textwidth}{!}{%
\begin{tabular}{|m{3cm}|m{8cm}|m{2cm}|m{2cm}|}
\hline
Category & Secondary Classification & AOIs & Buildings \\
\hline
Residential & Residential areas, dormitories & 527905 & 5162706 \\
\hline
Commercial & Shopping centers, markets, supermarkets & 142771 & 479150 \\
\hline
Public service & Education, medical care, venues and stations & 560775 & 1009976 \\
\hline
Industry & Factory & 46008 & 381129 \\
\hline
Office & Office building, government & 296943 & 875355 \\
\hline
\end{tabular}
}
\end{table}
\FloatBarrier 

\subsubsection*{Supplementary Table 3. Description of the attributes’ fields in data products.}
\small
\begin{longtable}{|m{2cm}|m{9.5cm}|m{3cm}|}
\hline
\textbf{Name} & \textbf{Description} & \textbf{Unit} \\
\hline
\endfirsthead

\hline
Name & Description & Unit \\
\hline
\endhead

\hline
\endfoot

\hline
\endlastfoot

Shape\_id & ID of the building & — \\
\hline
Rect\_Area & Minimum bounding rectangle area & m² \\
\hline
Aspect\_Rat & Building length-to-width ratio & — \\
\hline
Length & Building length & m \\
\hline
Width & Building width & m \\
\hline
Area\_Ratio & Ratio of contour area to ideal shape area & — \\
\hline
Shape\_A & Building base area & m² \\
\hline
Shape\_L & Building perimeter & m \\
\hline
NPI & Building compactness & — \\
\hline
MinCircle & Minimum Enclosing circle Area & m² \\
\hline
XzCity\_id & Administrative city Level & — \\
\hline
XzCity\_Nme & Administrative province & — \\
\hline
XzCity\_Nm & Administrative city & — \\
\hline
XzCity\_Nmx & Administrative county/district & — \\
\hline
Sum\_spbu & Total building base area & m² \\
\hline
Num\_bu & Number of buildings & — \\
\hline
Den\_bu & Building density & — \\
\hline
Dis\_Road & Distance to nearest secondary road & m \\
\hline
IfNearRoad & Is roadside building or not & — \\
\hline
CityFun & Distance to urban functional center & m \\
\hline
QHCITY\_id & Natural climatic subdivision & — \\
\hline
Pred\_h\_r & Building height & m \\
\hline
Floor & Building floor & — \\
\hline
Bu\_Area & Building area & m² \\
\hline
Block\_id & ID of the block on which the building is located & — \\
\hline
Block\_area & Block area & m² \\
\hline
Block\_leng & Block perimeter & m \\
\hline
StCity\_id & ID of the spatial city on which the building is located & — \\
\hline
ISWATER & Contain water bodies or not & — \\
\hline
NBL1 & Number of beauty salon types POIs & — \\
\hline
DBL1 & Density of beauty salon types POIs & Per m² \\
\hline
NBL2 & Number of transportation facility types POIs & — \\
\hline
DBL2 & Density of transportation facility types POIs & Per m² \\
\hline
NBL3 & Number of leisure and entertainment types POIs & — \\
\hline
DBL3 & Density of leisure and entertainment types POIs & Per m² \\
\hline
NBL4 & Number of company enterprise types POIs & — \\
\hline
DBL4 & Density of company enterprise types POIs & Per m² \\
\hline
NBL5 & Number of inlet and outlet types POIs & — \\
\hline
DBL5 & Density of inlet and outlet types POIs & Per m² \\
\hline
NBL6 & Number of medical treatment types POIs & — \\
\hline
DBL6 & Density of medical treatment types POIs & Per m² \\
\hline
NBL7 & Number of real estate types POIs & — \\
\hline
DBL7 & Density of real estate types POIs & Per m² \\
\hline
NBL8 & Number of governmental agencies types POIs & — \\
\hline
DBL8 & Density of governmental agencies types POIs & Per m² \\
\hline
NBL9 & Number of educational training types POIs & — \\
\hline
DBL9 & Density of educational training types POIs & Per m² \\
\hline
NBL10 & Number of cultural media types POIs & — \\
\hline
DBL10 & Density of cultural media types POIs & Per m² \\
\hline
NBL11 & Number of cultural media types POIs & — \\
\hline
DBL11 & Density of cultural media types POIs & Per m² \\
\hline
NBL12 & Number of car service types POIs & — \\
\hline
DBL12 & Density of car service types POIs & Per m² \\
\hline
NBL13 & Number of life service types POIs & — \\
\hline
DBL13 & Density of life service types POIs & Per m² \\
\hline
NBL14 & Number of restaurant types POIs & — \\
\hline
DBL14 & Density of restaurant types POIs & Per m² \\
\hline
NBL15 & Number of administrative landmark POIs & — \\
\hline
DBL15 & Density of administrative landmark POIs & Per m² \\
\hline
NBL16 & Number of shopping POIs & — \\
\hline
DBL16 & Density of shopping POIs & Per m² \\
\hline
NBL17 & Number of sports POIs & — \\
\hline
DBL17 & Density of sports POIs & Per m² \\
\hline
NBL18 & Number of hotel POIs & — \\
\hline
DBL18 & Density of hotel POIs & Per m² \\
\hline
NBL19 & Number of finance POIs & — \\
\hline
DBL19 & Density of finance POIs & Per m² \\
\hline
DBL & Total POIs density & Per m² \\
\hline
RBL1 & Proportion of beauty salon types POIs & — \\
\hline
RBL2 & Proportion of transportation facility types POIs & — \\
\hline
RBL3 & Proportion of leisure and entertainment types POIs & — \\
\hline
RBL4 & Proportion of company enterprise types POIs & — \\
\hline
RBL5 & Proportion of inlet and outlet types POIs & — \\
\hline
RBL6 & Proportion of medical treatment types POIs & — \\
\hline
RBL7 & Proportion of real estate types POIs & — \\
\hline
RBL8 & Proportion of governmental agencies types POIs & — \\
\hline
RBL9 & Proportion of educational training types POIs & — \\
\hline
RBL10 & Proportion of cultural media types POIs & — \\
\hline
RBL11 & Proportion of cultural media types POIs & — \\
\hline
RBL12 & Proportion of car service types POIs & — \\
\hline
RBL13 & Proportion of life service types POIs & — \\
\hline
RBL14 & Proportion of restaurant types POIs & — \\
\hline
RBL15 & Proportion of administrative landmark POIs & — \\
\hline
RBL16 & Proportion of shopping POIs & — \\
\hline
RBL17 & Proportion of sports POIs & — \\
\hline
RBL18 & Proportion of hotel POIs & — \\
\hline
RBL19 & Proportion of finance POIs & — \\
\hline
M\_Block & POIs diversity index & — \\
\hline
Sum\_Bu & Total building area & m² \\
\hline
Plot\_rat & Plot ratio & — \\
\hline
Mean\_H & Average Height of Buildings & m \\
\hline
Mean\_F & Average Floor of Buildings & — \\
\hline
type\_2023 & Building functions identified by AOI data & — \\
\hline
predict & Predicted building function & — \\
\hline
Age\_IS & Number of pixel identified by GAIA data & — \\
\hline
Age & Building age & — \\
\hline
StrVi\_100 & Name of existing street view observation POIsnts in the buffer & — \\
\hline
mFa\_13\_100 & The disorder score of Buildings with damaged facades in 2013 & — \\
\hline
mFa\_14\_100 & The disorder score of Buildings with damaged facades in 2014 & — \\
\hline
mFa\_15\_100 & The disorder score of Buildings with damaged facades in 2015 & — \\
\hline
mFa\_16\_100 & The disorder score of Buildings with damaged facades in 2016 & — \\
\hline
mFa\_17\_100 & The disorder score of Buildings with damaged facades in 2017 & — \\
\hline
mFa\_18\_100 & The disorder score of Buildings with damaged facades in 2018 & — \\
\hline
mFa\_19\_100 & The disorder score of Buildings with damaged facades in 2019 & — \\
\hline
mFa\_20\_100 & The disorder score of Buildings with damaged facades in 2020 & — \\
\hline
mFa\_21\_100 & The disorder score of Buildings with damaged facades in 2021 & — \\
\hline
mFa\_22\_100 & The disorder score of Buildings with damaged facades in 2022 & — \\
\hline
lBu\_13\_100 & The disorder score of Illegal/temporary buildings in 2013 & — \\
\hline
lBu\_14\_100 & The disorder score of Illegal/temporary buildings in 2014 & — \\
\hline
lBu\_15\_100 & The disorder score of Illegal/temporary buildings in 2015 & — \\
\hline
lBu\_16\_100 & The disorder score of Illegal/temporary buildings in 2016 & — \\
\hline
lBu\_17\_100 & The disorder score of Illegal/temporary buildings in 2017 & — \\
\hline
lBu\_18\_100 & The disorder score of Illegal/temporary buildings in 2018 & — \\
\hline
lBu\_19\_100 & The disorder score of Illegal/temporary buildings in 2019 & — \\
\hline
lBu\_20\_100 & The disorder score of Illegal/temporary buildings in 2020 & — \\
\hline
lBu\_21\_100 & The disorder score of Illegal/temporary buildings in2021 & — \\
\hline
lBu\_22\_100 & The disorder score of Illegal/temporary buildings in 2022 & — \\
\hline
lAd\_13\_100 & The disorder score of Graffiti/illegal advertisement in 2013 & — \\
\hline
lAd\_14\_100 & The disorder score of Graffiti/illegal advertisement in 2014 & — \\
\hline
lAd\_15\_100 & The disorder score of Graffiti/illegal advertisement in 2015 & — \\
\hline
lAd\_16\_100 & The disorder score of Graffiti/illegal advertisement in 2016 & — \\
\hline
lAd\_17\_100 & The disorder score of Graffiti/illegal advertisement in 2017 & — \\
\hline
lAd\_18\_100 & The disorder score of Graffiti/illegal advertisement in 2018 & — \\
\hline
lAd\_19\_100 & The disorder score of Graffiti/illegal advertisement in 2019 & — \\
\hline
lAd\_20\_100 & The disorder score of Graffiti/illegal advertisement in 2020 & — \\
\hline
lAd\_21\_100 & The disorder score of Graffiti/illegal advertisement in 2021 & — \\
\hline
lAd\_22\_100 & The disorder score of Graffiti/illegal advertisement in 2022 & — \\
\hline
kSt\_13\_100 & The disorder score of Stores with poor facades in 2013 & — \\
\hline
kSt\_14\_100 & The disorder score of Stores with poor facades in 2014 & — \\
\hline
kSt\_15\_100 & The disorder score of Stores with poor facades in 2015 & — \\
\hline
kSt\_16\_100 & The disorder score of Stores with poor facades in 2016 & — \\
\hline
kSt\_17\_100 & The disorder score of Stores with poor facades in 2017 & — \\
\hline
kSt\_18\_100 & The disorder score of Stores with poor facades in 2018 & — \\
\hline
kSt\_19\_100 & The disorder score of Stores with poor facades in 2019 & — \\
\hline
kSt\_20\_100 & The disorder score of Stores with poor facades in 2020 & — \\
\hline
kSt\_21\_100 & The disorder score of Stores with poor facades in 2021 & — \\
\hline
kSt\_22\_100 & The disorder score of Stores with poor facades in 2022 & — \\
\hline
kFa\_13\_100 & The disorder score of Buildings with unkempt facades in 2013 & — \\
\hline
kFa\_14\_100 & The disorder score of Buildings with unkempt facades in 2014 & — \\
\hline
kFa\_15\_100 & The disorder score of Buildings with unkempt facades in 2015 & — \\
\hline
kFa\_16\_100 & The disorder score of Buildings with unkempt facades in 2016 & — \\
\hline
kFa\_17\_100 & The disorder score of Buildings with unkempt facades in 2017 & — \\
\hline
kFa\_18\_100 & The disorder score of Buildings with unkempt facades in 2018 & — \\
\hline
kFa\_19\_100 & The disorder score of Buildings with unkempt facades in 2019 & — \\
\hline
kFa\_20\_100 & The disorder score of Buildings with unkempt facades in 2020 & — \\
\hline
kFa\_21\_100 & The disorder score of Buildings with unkempt facades in 2021 & — \\
\hline
kFa\_22\_100 & The disorder score of Buildings with unkempt facades in 2022 & — \\
\hline
oSi\_13\_100 & The disorder score of Stores with poor signboards in 2013 & — \\
\hline
oSi\_14\_100 & The disorder score of Stores with poor signboards in 2014 & — \\
\hline
oSi\_15\_100 & The disorder score of Stores with poor signboards in 2015 & — \\
\hline
oSi\_16\_100 & The disorder score of Stores with poor signboards in 2016 & — \\
\hline
oSi\_17\_100 & The disorder score of Stores with poor signboards in 2017 & — \\
\hline
oSi\_18\_100 & The disorder score of Stores with poor signboards in 2018 & — \\
\hline
oSi\_19\_100 & The disorder score of Stores with poor signboards in 2019 & — \\
\hline
oSi\_20\_100 & The disorder score of Stores with poor signboards in 2020 & — \\
\hline
oSi\_21\_100 & The disorder score of Stores with poor signboards in 2021 & — \\
\hline
oSi\_22\_100 & The disorder score of Stores with poor signboards in 2022 & — \\
\hline
aBu\_13\_100 & The disorder score of abandoned building in 2013 & — \\
\hline
aBu\_14\_100 & The disorder score of abandoned building in 2014 & — \\
\hline
aBu\_15\_100 & The disorder score of abandoned building in 2015 & — \\
\hline
aBu\_16\_100 & The disorder score of abandoned building in 2016 & — \\
\hline
aBu\_17\_100 & The disorder score of abandoned building in 2017 & — \\
\hline
aBu\_18\_100 & The disorder score of abandoned building in 2018 & — \\
\hline
aBu\_19\_100 & The disorder score of abandoned building in 2019 & — \\
\hline
aBu\_20\_100 & The disorder score of abandoned building in 2020 & — \\
\hline
aBu\_21\_100 & The disorder score of abandoned building in 2021 & — \\
\hline
aBu\_22\_100 & The disorder score of abandoned building in 2022 & — \\
\hline
cSt\_13\_100 & The disorder score of abandoned building in 2013 & — \\
\hline
cSt\_14\_100 & The disorder score of abandoned building in 2014 & — \\
\hline
cSt\_15\_100 & The disorder score of abandoned building in 2015 & — \\
\hline
cSt\_16\_100 & The disorder score of abandoned building in 2016 & — \\
\hline
cSt\_17\_100 & The disorder score of abandoned building in 2017 & — \\
\hline
cSt\_18\_100 & The disorder score of abandoned building in 2018 & — \\
\hline
cSt\_19\_100 & The disorder score of abandoned building in 2019 & — \\
\hline
cSt\_20\_100 & The disorder score of abandoned building in 2020 & — \\
\hline
cSt\_21\_100 & The disorder score of abandoned building in 2021 & — \\
\hline
cSt\_22\_100 & The disorder score of abandoned building in 2022 & — \\
\hline
BQ\_13\_100 & The disorder score of building quality in 2013 & — \\
\hline
BQ\_14\_100 & The disorder score of building quality in 2014 & — \\
\hline
BQ\_15\_100 & The disorder score of building quality in 2015 & — \\
\hline
BQ\_16\_100 & The disorder score of building quality in 2016 & — \\
\hline
BQ\_17\_100 & The disorder score of building quality in 2017 & — \\
\hline
BQ\_18\_100 & The disorder score of building quality in 2018 & — \\
\hline
BQ\_19\_100 & The disorder score of building quality in 2019 & — \\
\hline
BQ\_20\_100 & The disorder score of building quality in 2020 & — \\
\hline
BQ\_21\_100 & The disorder score of building quality in 2021 & — \\
\hline
BQ\_22\_100 & The disorder score of building quality in 2022 & — \\
\hline
BQ\_lst\_100 & Building quality problem index in latest year & — \\
\hline
BQ\_lsy\_100 & The latest year of street view image & — \\
\hline
\end{longtable}
\FloatBarrier 

\subsubsection*{Supplementary Table 4. Comparison of height partition model and combination model.}
\begin{table}[H]
\centering
\small
\resizebox{\textwidth}{!}{%
\begin{tabular}{|m{3.5cm}|m{5cm}|m{5cm}|m{2cm}|m{1.5cm}|m{1.5cm}|m{1.5cm}|}
\hline
\textbf{Method} & \textbf{Administrative level} & \textbf{Number of buildings} & \textbf{MAE} & \textbf{RMSE} & \textbf{R²} \\
\hline
Partition model A & Areas not in the urban system & 319249 & 2.5 & 5.6 & 0.84 \\
\hline
Partition model B & County-level cities & 682958 & 2.4 & 5.3 & 0.81 \\
\hline
Partition model C & Prefecture-level cities & 3011588 & 3.3 & 6.5 & 0.74 \\
\hline
Partition model D & Provincial capitals & 3375660 & 3.9 & 7.7 & 0.71 \\
\hline
Partition model E & Municipalities & 1448991 & 3.6 & 7.9 & 0.68 \\
\hline
Combination model & All data & 8838446 & 3.8 & 7.6 & 0.71 \\
\hline
\end{tabular}
}
\end{table}

\FloatBarrier 

\clearpage 
\renewcommand{\thefigure}{S\arabic{figure}}
\setcounter{figure}{0} 
\subsection*{Supplementary Tables}
\subsubsection*{Supplementary Figure 1. Time and area distribution of GES imagery for spatial cities.}
\begin{figure}[!htbp] 
\centering
\includegraphics[width=1\textwidth]{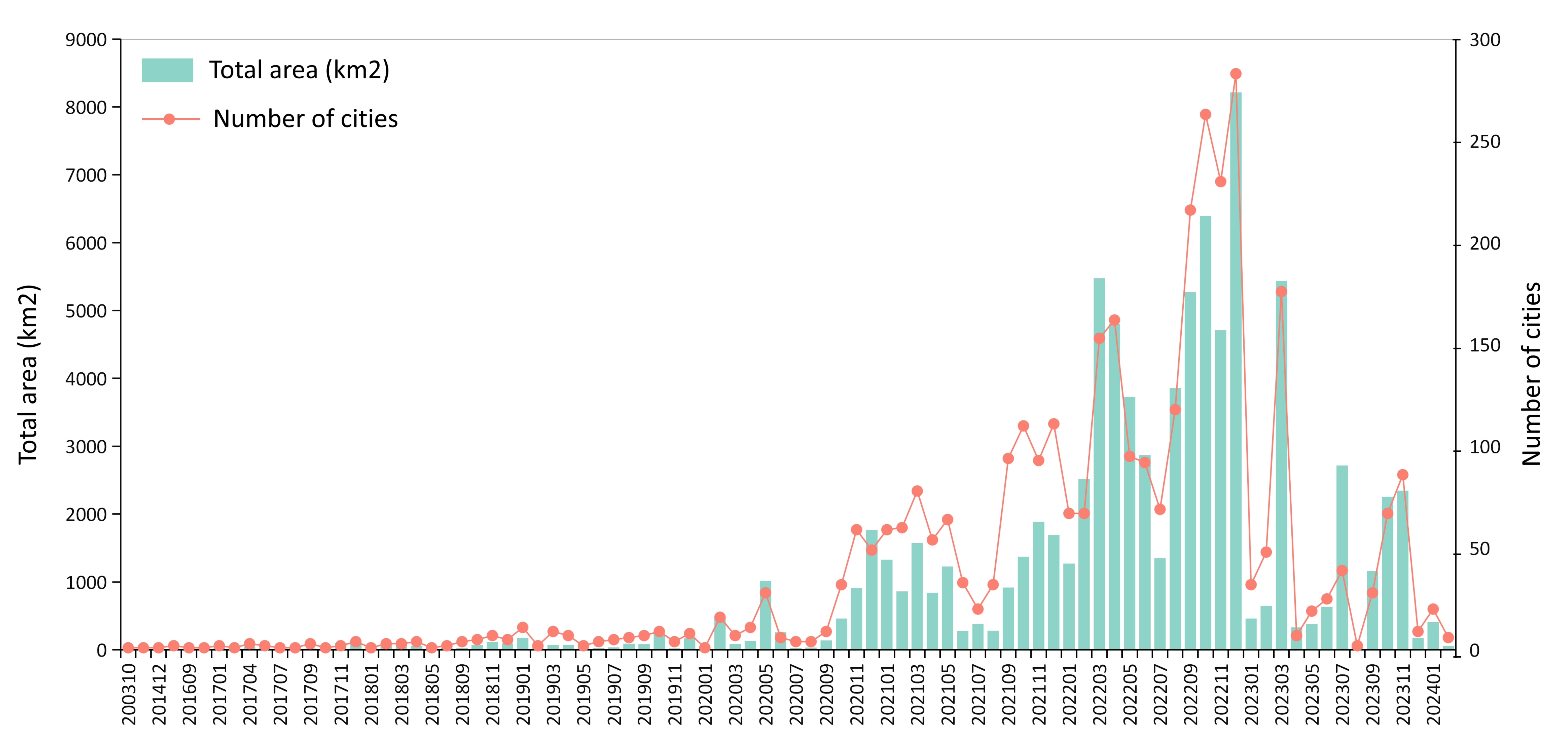}  
\caption{\textbf{Time and area distribution of GES imagery for spatial cities.}}
\label{fig:stream}
\end{figure}
\FloatBarrier 

\clearpage 
\subsubsection*{Supplementary Figure 2. Resampling process of the high-density urban areas with label enhancement.}
\begin{figure}[!htbp] 
\centering
\includegraphics[width=1\textwidth]{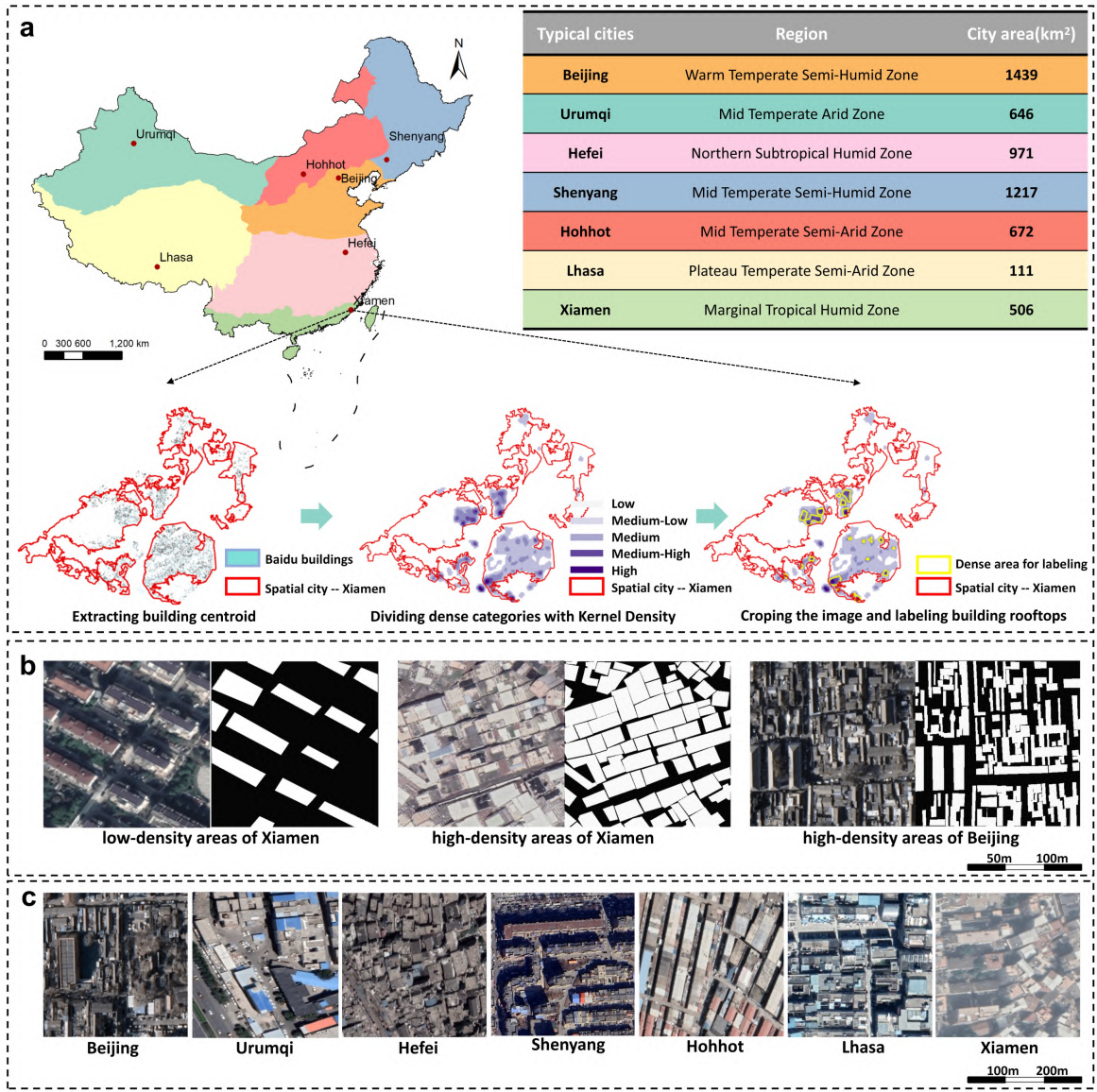}  
\caption{\textbf{Resampling process of the high-density urban areas with label enhancement.} \small{(a) Climate zones and labeled cities distribution. (b) Label examples for different density urban areas. (c) Different architectural types in typical cities.}}
\label{fig:stream}
\end{figure}
\FloatBarrier 

\clearpage 
\subsubsection*{Supplementary Figure 3. Statistics and preprocessing of Baidu building dataset.}
\begin{figure}[!htbp] 
\centering
\includegraphics[width=1\textwidth]{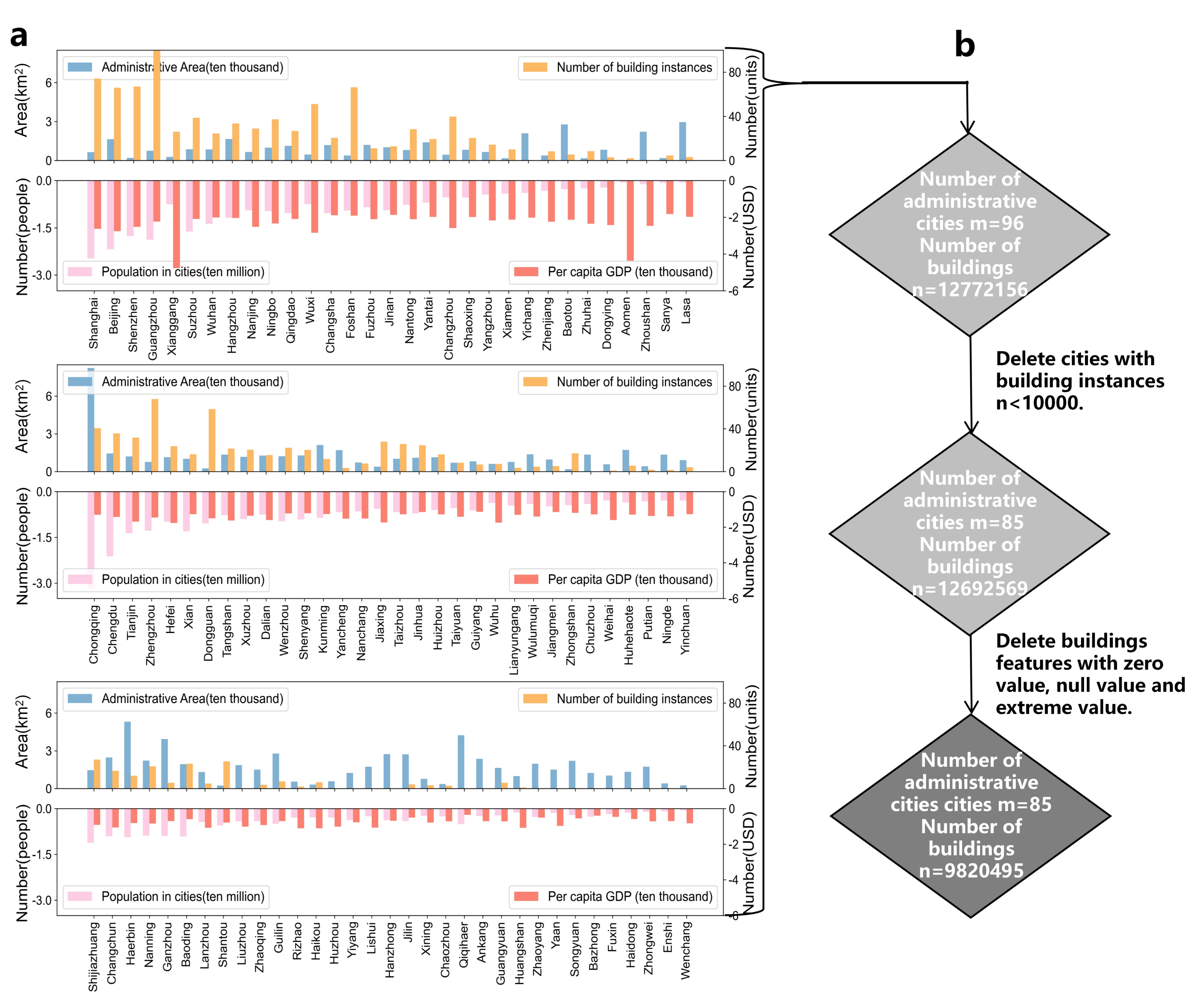}  
\caption{\textbf{Statistics and preprocessing of Baidu building dataset.} \small{(a) Statistics of cities in Baidu. (b) Data preprocessing. According to the spatial hierarchical sampling strategy, there are 62 C-level (prefecture), 28 D-level (provincial capital) and 6 E-level (municipality) cities in 96 Baidu cities. At the same time, according to the administrative boundaries of China, the building data of these cities also include the building data of county-level cities and even non-urban systems under the jurisdiction of the city, which ensures the coverage integrity of all types of urban building height samples. Bootstrap strategy with playback is used to ensure the balance of samples.}}
\label{fig:stream}
\end{figure}
\FloatBarrier 

\clearpage 
\subsubsection*{Supplementary Figure 4. The samples of building feature calculation.}
\begin{figure}[!htbp] 
\centering
\includegraphics[width=1\textwidth]{figures/figures_18.pdf}  
\caption{\textbf{The samples of building feature calculation.}}
\label{fig:stream}
\end{figure}
\FloatBarrier 

\clearpage 
\subsubsection*{\textbf{Supplementary Figure 5. Uncertainty expressed in the Relative Error and Absolute Error values of the aggregated method for partition model and combination model.} \small{(a) stands for the combined model that all regional buildings are trained together, and (b-f) stands for the model that is trained according to administrative levels.}}
\begin{figure}[!htbp] 
\centering
\includegraphics[width=1\textwidth]{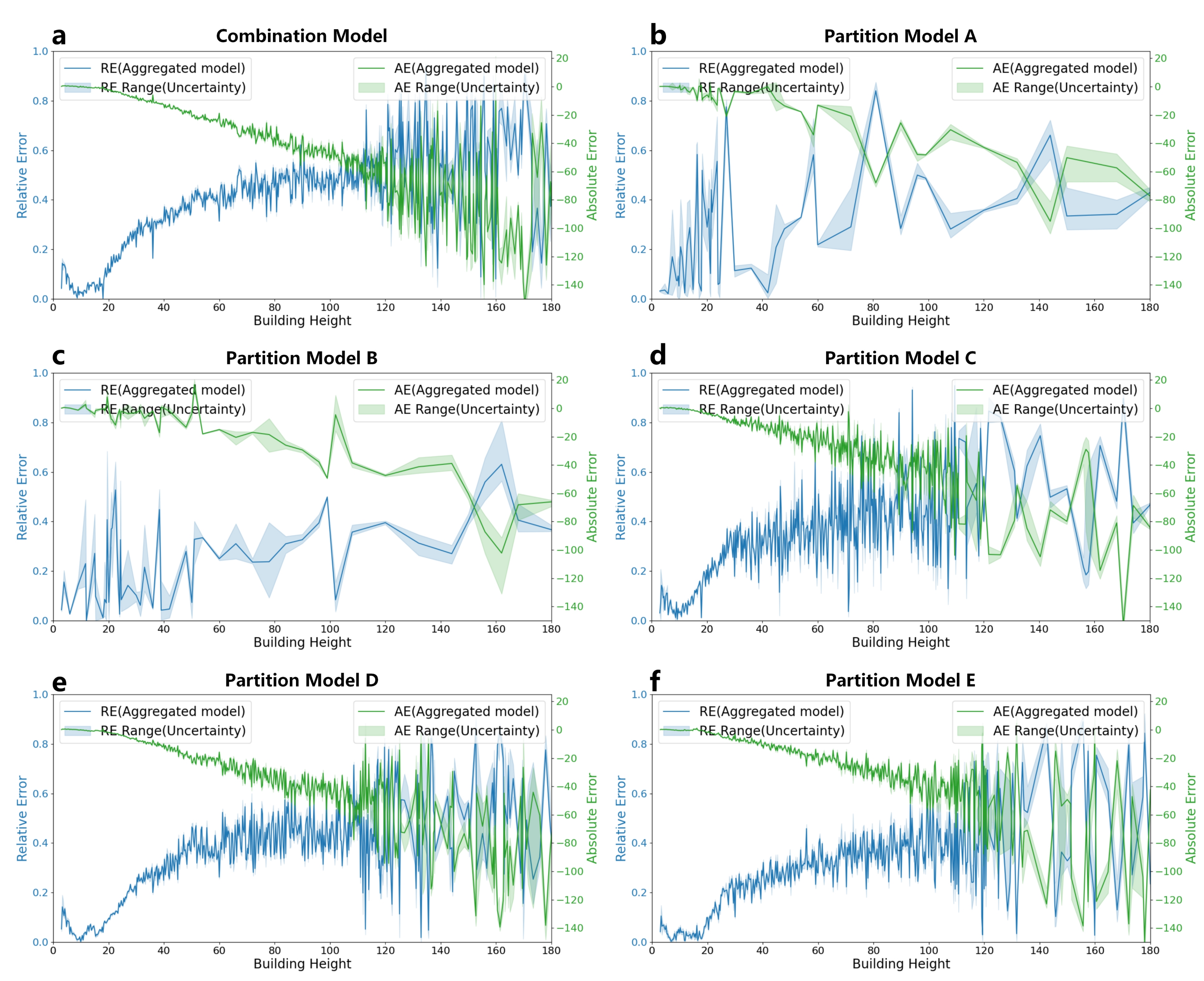}  
\caption{\textbf{Accuracy of building function model.}}
\label{fig:stream}
\end{figure}
\FloatBarrier 

\clearpage 
\subsubsection*{Supplementary Figure 6. Accuracy of building function model.}
\begin{figure}[!htbp] 
\centering
\includegraphics[width=1\textwidth]{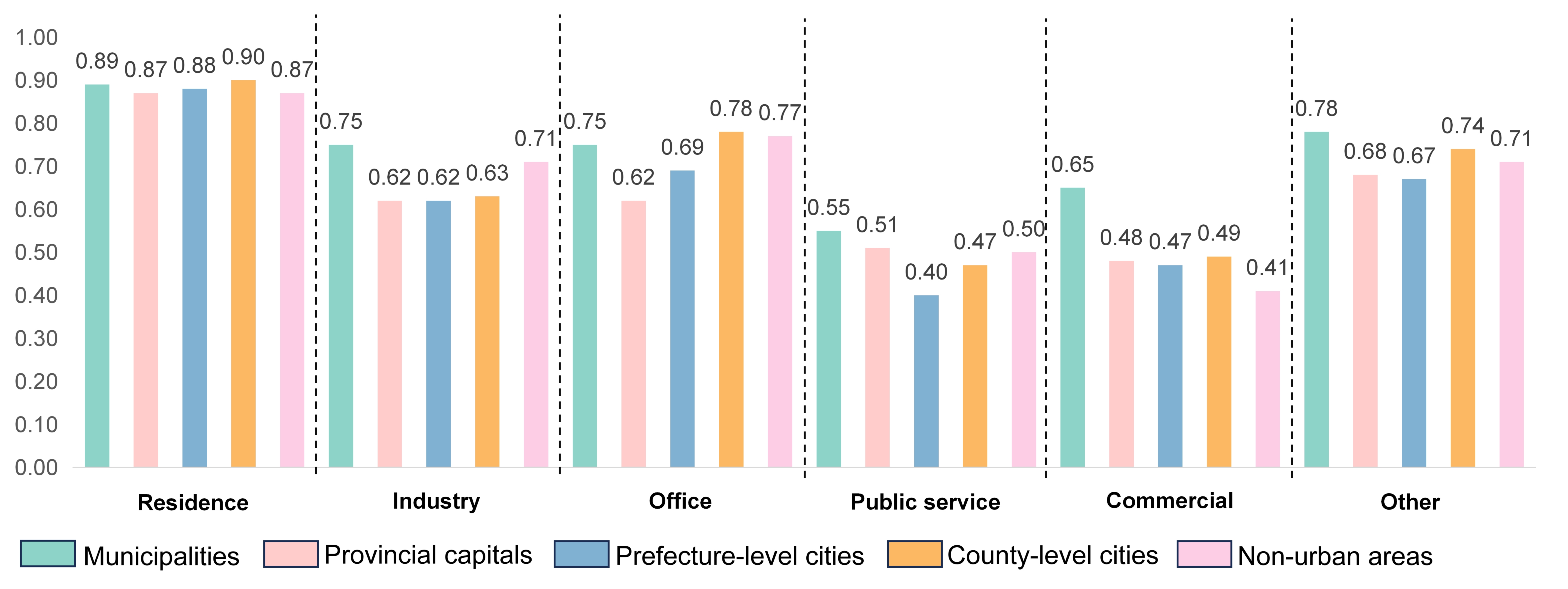}  
\caption{\textbf{Accuracy of building function model.}}
\label{fig:stream}
\end{figure}
\FloatBarrier 


\begin{thebibliography}{99}

\bibitem[Akiyama et al., 2019]{AY2017} Akiyama, Y. et al. Estimating the spatial distribution of vacant houses using public municipal data. In: Int. Conf. Geogr. Inf. Sci. Springer, Cham: 165-183 (2019).

\bibitem[Alobeid et al., 2009]{AJK2009} Alobeid, A., Jacobsen, K. \& Heipke, C. Building height estimation in urban areas from very high resolution satellite stereo images. In: ISPRS Hannover Workshop. 5, 2–5 (2009).

\bibitem[Biljecki et al., 2022]{BFCY2022} Biljecki, F. \& Chow, Y. S. Global building morphology indicators. Comput. Environ. Urban Syst. 95, 101809 (2022).

\bibitem[Biljecki et al., 2023]{BC2023} Biljecki, F., Chow, Y. S. \& Lee, K. Quality of crowdsourced geospatial building information: A global assessment of OpenStreetMap attributes. Build. Environ. 237, 110295 (2023).

\bibitem[Biljecki et al., 2017]{BL2017} Biljecki, F., Ledoux, H. \& Stoter, J. Generating 3d city models without elevation data. Comput. Environ. Urban Syst. 64, 1–18 (2017).

\bibitem[BITC, 2021]{BITC2021} BITC. Science Data Bank. DOI: 10.11922/sciencedb.00620. (2021).

\bibitem[Breiman, 1996]{BR1996} Breiman, L. Bagging predictors. Mach. Learn. 24, 123-140 (1996).

\bibitem[Cao et al., 2021]{CQ2021} Cao, Q. et al. The effects of 2D and 3D building morphology on urban environments: A multi-scale analysis in the Beijing metropolitan region. Build. Environ. 192, 107635 (2021).

\bibitem[Cao \& Huang, 2021]{CH2021} Cao, Y. \& Huang, X. A deep learning method for building height estimation using high-resolution multi-view imagery over urban areas: A case study of 42 Chinese cities. Remote Sens. Environ. 264, 112590 (2021).

\bibitem[Cao \& Weng, 2024]{CQ2024} Cao, Y. \& Weng, Q. A deep learning-based super-resolution method for building height estimation at 2.5 m spatial resolution in the Northern Hemisphere. Remote Sens. Environ. 310, 114241 (2024).

\bibitem[Che et al., 2024]{CY2024} Che, Y. et al. 3D-GloBFP: the first global three-dimensional building footprint dataset. Earth Syst. Sci. Data Discuss. 1-28 (2024).

\bibitem[Chen et al., 2022]{CD2022} Chen, D. et al. A hierarchical approach for fine-grained urban villages recognition fusing remote and social sensing data. Int. J. Appl. Earth Obs. Geoinf. 106, 102661 (2022).

\bibitem[Chen et al., 2023a]{CJ2023} Chen, J. et al. Measuring physical disorder in urban street spaces: a large-scale analysis using street view images and deep learning. Ann. Am. Assoc. Geogr. 113, 469-487 (2023a).

\bibitem[Chen \& Guestrin, 2016]{CG2016} Chen, T. \& Guestrin, C. Xgboost: A scalable tree boosting system. In: Proc. 22nd ACM SIGKDD Int. Conf. Knowl. Discov. Data Min. (2016).

\bibitem[Chen et al., 2023b]{CW2023b} Chen, W. et al. Large-scale urban building function mapping by integrating multi-source web-based geospatial data. Geo-spatial Inf. Sci. 1-15 (2023b).

\bibitem[Chen et al., 2024a]{CY2024a} Chen, Y. et al. Refining urban morphology: An explainable machine learning method for estimating footprint-level building height. Sustain. Cities Soc. 105635 (2024a).

\bibitem[Chen et al., 2024b]{CZ2024b} Chen, Z. et al. City-scale solar PV potential estimation on 3D buildings using multi-source RS data: A case study in Wuhan, China. Appl. Energy 359, 122720 (2024b).

\bibitem[Chhetri et al., 2013]{CP2013} Chhetri, P. et al. Mapping urban residential density patterns: Compact city model in Melbourne, Australia. City Cult. Soc. 4(2), 77-85 (2013).

\bibitem[Dempsey, 2010]{DN2010} Dempsey, N. Revisiting the Compact City, Dempsey, N., (NOT FINAL PROOF) published in BUILT ENVIRONMENT VOL 36 NO 1, pp. 5-8. Environ. 36(1), 5-8 (2010).

\bibitem[Deng et al., 2023]{DX2023} Deng, X. et al. Characterizing urban densification and quantifying its effects on urban thermal environments and human thermal comfort. Landsc. Urban Plan. 237, 104803 (2023).

\bibitem[Deng et al., 2022]{DY2022} Deng, Y. et al. Identify urban building functions with multisource data: A case study in Guangzhou, China. Int. J. Geogr. Inf. Sci. 36(10), 2060-2085 (2022).

\bibitem[Dovey \& Pafka, 2014]{DK2014} Dovey, K. \& Pafka, E. The urban density assemblage: Modelling multiple measures. Urban Des. Int. 19, 66-76 (2014).

\bibitem[Du et al., 2018]{DS2018} Du, S. et al. Exploring the changes of building vulnerability and its implications for urban flood risk in Shanghai. Sustainability 10(6), 1961 (2018).

\bibitem[Efron \& Tibshirani, 1993]{EB1993} Efron, B. \& Tibshirani, R. J. An introduction to the bootstrap. (1993).

\bibitem[Fang et al., 2021]{FF2021} Fang, F. et al. Dataset of building instances in typical Chinese cities [DB/OL]. Sci. Data Bank. Available at: https://doi.org/10.11922/sciencedb.00620 (2021).

\bibitem[Ferreira \& Batey, 2011]{FA2011} Ferreira, A. \& Batey, P. On why planning should not reinforce self-reinforcing trends: a cautionary analysis of the compact-city proposal applied to large cities. Environ. Plann. B Plann. Des. 38(2), 231-247 (2011).

\bibitem[Friedman et al., 2000]{FJ2000} Friedman, J., Hastie, T. \& Tibshirani, R. Additive logistic regression: a statistical view of boosting (with discussion and a rejoinder by the authors). Ann. Stat. 28(2), 337-407 (2000).

\bibitem[Gao et al., 2021]{GF2021} Gao, F. et al. Integrating the Eigendecomposition Approach and k-Means Clustering for Inferring Building Functions with Location-Based Social Media Data. ISPRS Int. J. Geo-Inf. 10(12), 834 (2021).

\bibitem[GlobalMLBuildingFootprints, 2024]{GB2024} GlobalMLBuildingFootprints. Available at: https://github.com/microsoft/GlobalMLBuildingFootprints (accessed 16 Aug 2024).

\bibitem[Gong et al., 2020]{GP2020} Gong, P. et al. Annual maps of global artificial impervious area (GAIA) between 1985 and 2018. Remote Sens. Environ. 236, 111510 (2020).

\bibitem[Gröger \& Plümer, 2012]{GG2012} Gröger, G. \& Plümer, L. CityGML–Interoperable semantic 3D city models. ISPRS J. Photogramm. Remote Sens. 71, 12-33 (2012).

\bibitem[Gu et al., 2023]{GY2023} Gu, Y. et al. Generative design method of building group based on AIP (Aging in Place) assessment: The case of dense urban renewal districts in Hong Kong. In World Congress of Architects 479-495 (Springer International Publishing, 2023).

\bibitem[Herfort et al., 2023]{HB2023} Herfort, B. et al. A spatio-temporal analysis investigating completeness and inequalities of global urban building data in OpenStreetMap. Nat. Commun. 14(1), 3985 (2023).

\bibitem[Hoffmann et al., 2023]{HE2023} Hoffmann, E. J., Abdulahhad, K. \& Zhu, X. X. Using social media images for building function classification. Cities 133, 104107 (2023).

\bibitem[Huang et al., 2022]{HX2022} Huang, X. et al. Urban Building Classification (UBC) – A Dataset for Individual Building Detection and Classification from Satellite Imagery. In: Proc. IEEE/CVF Conf. Comput. Vis. Pattern Recognit. (2022).

\bibitem[Hui et al., 2018]{HJ2018} Hui, J. et al. Effective building extraction from high-resolution remote sensing images with multitask driven deep neural network. IEEE Geosci. Remote Sens. Lett. 16(5), 786-790 (2018).

\bibitem[Kadhim \& Mourshed, 2017]{KN2017} Kadhim, N. \& Mourshed, M. A shadow-overlapping algorithm for estimating building heights from VHR satellite images. IEEE Geosci. Remote Sens. Lett. 15(1), 8-12 (2017).

\bibitem[Kang et al., 2018]{KJ2018} Kang, J. et al. Building instance classification using street view images. ISPRS J. Photogramm. Remote Sens. 145, 44-59 (2018).

\bibitem[Koks \& Haer, 2020]{KE2020} Koks, E. E. \& Haer, T. A comprehensive assessment of climate change impacts on global flood risk. Int. J. Disaster Risk Reduct. 47, 101518 (2020).

\bibitem[Kolbe, 2009]{KT2009} Kolbe, T. H. Representing and exchanging 3D city models with CityGML. In: 3D Geo-Inf. Sci. 15-31 (2009).

\bibitem[Kong et al., 2024]{KB2024} Kong, B. et al. A graph-based neural network approach to integrate multi-source data for urban building function classification. Comput. Environ. Urban Syst. 110, 102094 (2024).

\bibitem[Lemonsu et al., 2015]{LA2015} Lemonsu, A. et al. Vulnerability to heat waves: Impact of urban expansion scenarios on urban heat island and heat stress in Paris (France). Urban Clim. 14, 586-605 (2015).

\bibitem[Li et al., 2023]{LH2023} Li, H. et al. Semi-supervised Learning from Street-View Images and OpenStreetMap for Automatic Building Height Estimation. arXiv (2023). Available at: https://arxiv.org/abs/2307.02574.

\bibitem[Li et al., 2018]{LJ2018} Li, J., Long, Y. \& Dang, A. Live-Work-Play Centers of Chinese cities: Identification and temporal evolution with emerging data. Comput. Environ. Urban Syst. 71, 58-66 (2018).

\bibitem[Li et al., 2024a]{LK2024} Li, K. \& Zeng, H. Multidisciplinary parameters for characterizing the 3D urban morphology: An overview based on the relational perspective. Sustain. Cities Soc. 105364 (2024a).

\bibitem[Li et al., 2020]{LM2020} Li, M. et al. Continental-scale mapping and analysis of 3D building structure. Remote Sens. Environ. 245, 111859 (2020).

\bibitem[Li et al., 2024b]{LQ2024} Li, Q. et al. A Review of Building Extraction From Remote Sensing Imagery: Geometrical Structures and Semantic Attributes. IEEE Trans. Geosci. Remote Sensing 62, 1–15 (2024b).

\bibitem[Li et al., 2024c]{LW2024} Li, W. et al. Rethinking the country-level percentage of population residing in urban area with a global harmonized urban definition. iScience (2024b).

\bibitem[Li et al., 2020]{LX2020} Li, X. et al. Mapping global urban boundaries from the global artificial impervious area (GAIA) data. Environ. Res. Lett. 15, 094044 (2020).

\bibitem[Li et al., 2024d]{LY2024} Li, Y. et al. Protocol for assessing neighborhood physical disorder using the YOLOv8 deep learning model. STAR protocols, 5(1), 102778 (2024d).

\bibitem[Li et al., 2023]{LY2023} Li, Y. et al. Greening the concrete jungle: Unveiling the co-mitigation of greenspace configuration on PM2.5 and land surface temperature with explanatory machine learning. Urban For. Urban Green. 128086 (2023).

\bibitem[Lin et al., 2021]{LA2021} Lin, A. et al. Identifying urban building function by integrating remote sensing imagery and POIs data. IEEE J. Sel. Top. Appl. Earth Obs. Remote Sens. (2021).

\bibitem[Liu et al., 2020a]{LC2020a} Liu, C.-J. et al. IM2ELEVATION: Building Height Estimation from Single-View Aerial Imagery. Remote Sens. 12(17), 2719 (2020a).

\bibitem[Liu et al., 2021]{LM2021} Liu, M. et al. High-resolution mapping of mainland China’s urban floor area. Landsc. Urban Plann. 214, 104187 (2021).

\bibitem[Liu et al., 2022]{LW2022} Liu, W. et al. Associatively segmenting semantics and estimating height from monocular remote-sensing imagery. IEEE Trans. Geosci. Remote Sensing 60, 1-17 (2022).

\bibitem[Liu et al., 2024]{LX2024} Liu, X. et al. Global Mapping of Three-Dimensional (3D) Urban Structures Reveals Escalating Utilization in the Vertical Dimension and Pronounced Building Space Inequality. Engineering (2024).

\bibitem[Liu et al., 2020b]{LX2020b} Liu, X. et al. High-spatiotemporal-resolution mapping of global urban change from 1985 to 2015. Nat. Sustain. 3, 564-570 (2020b).

\bibitem[Liu et al., 2023]{LZ2023} Liu, Z. et al. CBRA: The first multi-annual (2016–2021) and high-resolution (2.5 m) building rooftop area dataset in China derived with Super-resolution Segmentation from Sentinel-2 imagery. Earth Syst. Sci. Data Discuss. 1-40 (2023).

\bibitem[Mao et al., 2022]{MY2022} Mao, Y. et al. Beyond single receptive field: A receptive field fusion-and-stratification network for airborne laser scanning POIsnt cloud classification. ISPRS J. Photogramm. Remote Sens. 188, 45-61 (2022).

\bibitem[Ma et al., 2020]{ML2020} Ma, S.., Long, Y. Functional urban area delineations of cities on the Chinese mainland using massive Didi ride-hailing records. Cities, 97, p.102532 (2020).

\bibitem[Ma et al., 2019]{ML2019} Ma, S.., Long, Y. Identifying Spatial Cities in China at the Community Scale. J. Urban Reg. Plan, 11(1):37-50 (2019). 

\bibitem[Mao et al., 2023]{MY2023} Mao, Y. et al. Elevation estimation-driven building 3-D reconstruction from single-view remote sensing imagery. IEEE Trans. Geosci. Remote Sens. 61, 1-18 (2023).

\bibitem[Nguyen et al., 2023]{NM2023} Nguyen, M. H., Nguyen, T. A. \& Ly, H. B. Ensemble XGBoost schemes for improved compressive strength prediction of UHPC. Structures 57, 105062 (2023).

\bibitem[Ogawa et al., 2023]{OY2023} Ogawa, Y. et al. Deep learning approach for classifying the built year and structure of individual buildings by automatically linking street view images and GIS building data. IEEE J. Sel. Top. Appl. Earth Obs. Remote Sens. 16, 1740-1755 (2023).

\bibitem[Paprotny et al., 2020]{PD2020} Paprotny, D. et al. HANZE: A pan-European database of exposure to natural hazards and damaging historical floods since 1870. Earth Syst. Sci. Data 12(2), 981-1005 (2020).

\bibitem[Park \& Guldmann, 2019]{PY2019} Park, Y. \& Guldmann, J.-M. Creating 3D city models with building footprints and LIDAR POIsnt cloud classification: A machine learning approach. Comput. Environ. Urban Syst. 75, 76–89 (2019).

\bibitem[Pinho et al., 2023]{PM2023} Pinho, M. G. M. et al. The quality of OpenStreetMap food-related POIsnt-of-interest data for use in epidemiological research. Health Place 83, 103075 (2023).

\bibitem[Ramalingam \& Kumar, 2023]{RS2023} Ramalingam, S. P. \& Kumar, V. Automatizing the generation of building usage maps from geotagged street view images using deep learning. Build. Environ. 235, 110215 (2023).

\bibitem[Rashid, 2018]{RM2018} Rashid, M. Geometry of urban layouts: A Global Comparative Study. (2018). Springer.

\bibitem[Seto \& Shepherd, 2009]{SK2009} Seto, K. C. \& Shepherd, J. M. Global urban land-use trends and climate impacts. Curr. Opin. Environ. Sustain. 1(1), 89-95 (2009).

\bibitem[Shi et al., 2024]{SQ2024} Shi, Q. et al. The Last Puzzle of Global Building Footprints—Mapping 280 Million Buildings in East Asia Based on VHR Images. J. Remote Sens. 4, 0138 (2024).

\bibitem[Sirko et al., 2021]{SW2021} Sirko, W. et al. Continental-scale building detection from high resolution satellite imagery. arXiv (2021). Available at: https://arxiv.org/abs/2107.12283.

\bibitem[Sun et al., 2019]{SG2019} Sun, G. et al. Combinational shadow index for building shadow extraction in urban areas from Sentinel-2A MSI imagery. Int. J. Appl. Earth Obs. Geoinf. 78, 53-65 (2019).

\bibitem[Sun et al., 2022]{SM2022} Sun, M. et al. Understanding architecture age and style through deep learning. Cities 128, 103787 (2022).

\bibitem[Sun et al., 2024]{SX2024} Sun, X. et al. GABLE: A first fine-grained 3D building model of China on a national scale from very high resolution satellite imagery. Remote Sens. Environ. 305, 114057 (2024).

\bibitem[Taubenböck et al., 2018]{TH2018} Taubenböck, H. et al. The morphology of the Arrival City-A global categorization based on literature surveys and remotely sensed data. Appl. Geogr. 92, 150-167 (2018).

\bibitem[Tomer et al., n.d.]{TA2024} Tomer, A., Kane, J. W. \& Fishbane, L. Connecting people by proximity: A better way to plan metro areas.

\bibitem[Tusting et al., 2019]{TL2019} Tusting, L. S. et al. Mapping changes in housing in sub-Saharan Africa from 2000 to 2015. Nature 568(7752), 391-394 (2019).

\bibitem[Wang et al., 2024]{WH2024} Wang, L., Hou, C., Zhang, Y. \& He, J. Measuring solar radiation and spatio-temporal distribution in different street network direction through solar trajectories and street view images. Int. J. Appl. Earth Obs. Geoinf. 132, 104058 (2024).

\bibitem[Wu et al., 2023]{WW2023} Wu, W. B. et al. A first Chinese building height estimate at 10 m resolution (CNBH-10 m) using multi-source earth observations and machine learning. Remote Sens. Environ. 291, 113578 (2023).

\bibitem[Wu et al., 2024]{WX2024} Wu, X. et al. Exploring urban building space provision and inequality: A three-dimensional perspective on Chinese cities toward sustainable development goals. Sustain. Cities Soc. 102, 105202 (2024).

\bibitem[Yao et al., 2018]{YN2018} Yao, N. et al. Drought evolution, severity and trends in mainland China over 1961–2013. Sci. Total Environ. 616, 73-89 (2018).

\bibitem[Yin et al., 2024]{YL2024} Yin, L. et al. Rethinking demolition plans to fight neighborhood blight in shrinking cities: Applying agent-based policy simulations. Cities 150, 105035 (2024).

\bibitem[Zhang et al., 2023]{ZX2023} Zhang, X. et al. Inferring building function: A novel geo-aware neural network supporting building-level function classification. Sustain. Cities Soc. 89, 104349 (2023).

\bibitem[Zhang et al., 2022a]{ZH2022} Zhang, Y., Zhang, Q., Zhao, Y., Deng, Y. \& Zheng, H. Urban spatial risk prediction and optimization analysis of POIs based on deep learning from the perspective of an epidemic. Int. J. Appl. Earth Obs. Geoinf. 112, 102942 (2022a).

\bibitem[Zhang et al., 2022b]{ZZ2022} Zhang, Z. et al. Vectorized rooftop area data for 90 cities in China. Sci. Data 9, 66 (2022b).

\bibitem[Zhang et al., 2022c]{ZY2022} Zhang, Y. et al. Artificial intelligence prediction of urban spatial risk factors from an epidemic perspective. In Int. Conf. Comput. Des. Robot. Fabr. 209–222 (Springer Nature Singapore, 2022c).

\bibitem[Zheng et al., 2024]{ZY2024} Zheng, Y. et al. Identifying building function using multisource data: A case study of China's three major urban agglomerations. Sustain. Cities Soc. 108, 105498 (2024).

\bibitem[Zhou et al., 2017]{ZY2017} Zhou, Y. et al. A global record of annual urban dynamics (1992–2013) from nighttime lights. Remote Sens. Environ. 193, 1-10 (2017).


\end{thebibliography}
\end{document}